\documentclass[a4paper,fleqn]{cas-sc}

\usepackage[authoryear,longnamesfirst]{natbib}
\usepackage{xcolor}
\usepackage[utf8]{inputenc}
\usepackage{booktabs}
\usepackage{amssymb}
\usepackage{amsmath}
\usepackage{multirow}
\usepackage{pifont}
\usepackage{lineno}
\usepackage{tabularx}
\usepackage{graphicx}
\usepackage{float}
\usepackage{placeins}
\usepackage{flafter}
\usepackage{array}
\usepackage{longtable}
\AtBeginDocument{%
	\hypersetup{colorlinks=true, linkcolor=blue, citecolor=blue, urlcolor=blue}%
}

\begin{document}
\let\WriteBookmarks\relax
\def\floatpagepagefraction{1}
\def\textpagefraction{.001}
\setcounter{topnumber}{5}
\setcounter{bottomnumber}{5}
\setcounter{totalnumber}{10}
\renewcommand{\topfraction}{0.95}
\renewcommand{\bottomfraction}{0.95}
\renewcommand{\textfraction}{0.05}
\renewcommand{\floatpagefraction}{0.80}
\shorttitle{Edge-Aware and Content-Adaptive Infrared Gas Leak Detection}
\shortauthors{Li et~al.}

\title [mode = title]{Edge-Aware and Content-Adaptive Infrared Gas Leak Detection for Industrial Safety Monitoring}

\author[1]{Dongsheng Li}
\author[2]{Tianli Ma}
\author[2]{Siling Wang}
\author[3]{Beibei Duan}
\author[2]{Song Gao}
\cormark[1]
\ead{gaos@xatu.edu.cn}

\affiliation[1]{organization={School of Mechatronic Engineering, Xi'an Technological University},
            city={Xi'an},
            postcode={710021},
            state={Shaanxi},
            country={China}}

\affiliation[2]{organization={School of Electronic Information Engineering, Xi'an Technological University},
            city={Xi'an},
            postcode={710021},
            state={Shaanxi},
            country={China}}

\affiliation[3]{organization={Shaanxi Shanhua Coal Chemical Co., Ltd.},
            city={Weinan},
            postcode={714104},
            state={Shaanxi},
            country={China}}

\cortext[1]{Corresponding author. Full postal address: School of Electronic Information Engineering, Xi'an Technological University, Xi'an, 710021, Shaanxi, China.}

	\begin{abstract}
Infrared gas leak detection is important for industrial safety and environmental monitoring, but automatic detection remains challenging because gas plumes are often faint, small, semi-transparent, and weakly bounded. This study proposes an Edge-Aware and Content-Adaptive Feature Fusion Detector (ECAF-Det) for infrared gas leak detection in weak-plume and cluttered thermal scenes.
{The main methodological contributions of ECAF-Det comprise
	three task-oriented components. A local--global feature enhancement block preserves fine boundary cues and long-range plume continuity. A multi-scale edge perception module transforms directional-gradient and Gabor-response cues into hierarchical boundary-sensitive structural priors. A content-adaptive sparse routing path aggregation network dynamically regulates multi-scale feature propagation and limits the contribution of less informative cross-scale responses.}
{Experiments on the IIG dataset show that ECAF-Det improves overall and small-plume detection while maintaining moderate computational complexity. On this dataset, ECAF-Det achieves an average precision (AP) of 29.8\%, an AP at an IoU threshold of 0.5 (AP$_{50}$) of 84.3\%, and a small-object AP of 25.3\%. Compared with the Real-Time Detection Transformer with a ResNet-18 backbone (RT-DETR-R18), these values represent improvements of 3.0, 6.5, and 5.4 percentage points, respectively. The model requires 43.7 giga floating-point operations (GFLOPs) and 14.3 M parameters.}
On the LangGas dataset, ECAF-Det achieves an AP of 36.3\% and an AP$_{50}$ of 68.5\%. The AI contribution lies in edge-aware representation learning and content-adaptive sparse feature routing for weak infrared plume perception. The engineering application is automated infrared gas leak detection for industrial safety monitoring, early warning, and remote inspection.
\end{abstract}

\begin{keywords}
Infrared gas leak detection \sep Engineering artificial intelligence \sep Industrial safety monitoring \sep Weak-plume detection \sep Edge-aware representation \sep Content-adaptive feature fusion
\end{keywords}

\maketitle



\section{Introduction}
Gas leakage from petrochemical plants, natural gas facilities, and pipeline networks remains a major challenge for industrial safety and environmental protection. As process industries continue to expand, increasingly dense equipment layouts and extensive pipeline systems increase the risk that small leaks may remain undetected until they escalate into serious incidents~\citep{xu2023outlook,lu2020leakage}. Gas leaks can trigger fires and explosions and may also release toxic, flammable, or environmentally harmful substances into the surrounding environment~\citep{zuo2023leak,bonvicini2015quantitative,kopbayev2022gas}. These releases can threaten workers, nearby communities, ecosystems, and public health. Therefore, timely and reliable gas leak detection is essential for accident prevention, emergency response, and environmental risk mitigation.

{In practical industrial facilities, gas-leak monitoring still relies on manual inspection and fixed gas sensors, while acoustic, pressure-wave, and infrared sensing provide complementary approaches for leakage monitoring. Fixed gas sensors are widely used but provide limited spatial coverage and can be affected by environmental conditions~\citep{wang2023review}. Acoustic monitoring and negative-pressure-wave methods have also been investigated for pipeline leakage detection~\citep{quy2022pipeline,han2023novel}, although their performance can depend on background interference, sensor placement, pipeline structure, and signal propagation conditions. Infrared (IR) thermography is particularly attractive for remote monitoring because it provides spatially resolved plume information without direct contact with the leakage source~\citep{strahl2021methane}. Nevertheless, in practical industrial scenes, infrared gas plumes may exhibit weak and unstable contrast, especially for small leaks, low concentrations, long imaging distances, and cluttered thermal backgrounds. These characteristics motivate the development of automated visual detection methods specifically tailored to infrared gas plumes.}

{Deep-learning-based detectors provide a useful foundation for automated infrared gas leak detection. However, generic detection architectures are not specifically designed for the distinctive visual characteristics of gas plumes~\citep{zhang2025m4net,zhong2024hierarchical,wang2026miigan}. Unlike rigid objects with stable contours and textures, infrared gas plumes often exhibit weak contrast, diffuse and unstable boundaries, semi-transparency, and large variations in spatial scale. Weak plume responses can therefore be attenuated during hierarchical feature extraction, while fixed multi-scale fusion may propagate less informative thermal-background responses. Consequently, effective infrared gas detection requires task-oriented representation mechanisms that preserve weak local structures, capture long-range plume continuity, and adapt cross-scale feature propagation while maintaining moderate computational cost.}

From an engineering application perspective, the objective of infrared gas
leak detection is not only to improve visual detection accuracy, but also to
reduce missed alarms for weak early-stage leaks and provide spatial evidence
for operator decision-making. In practical monitoring scenarios, a missed weak plume may delay emergency response, whereas excessive false alarms may increase inspection burden and reduce operator trust in the monitoring system. Therefore, gas-leak detection models should be evaluated not only by general object-detection metrics, but also by weak-plume recall, false-alarm tendency, computational latency, and small-leak detectability. These considerations motivate a deployment-oriented detector that improves faint-plume perception while maintaining practical computational feasibility.
\begin{figure}
	\centering
	\includegraphics[width=\textwidth]{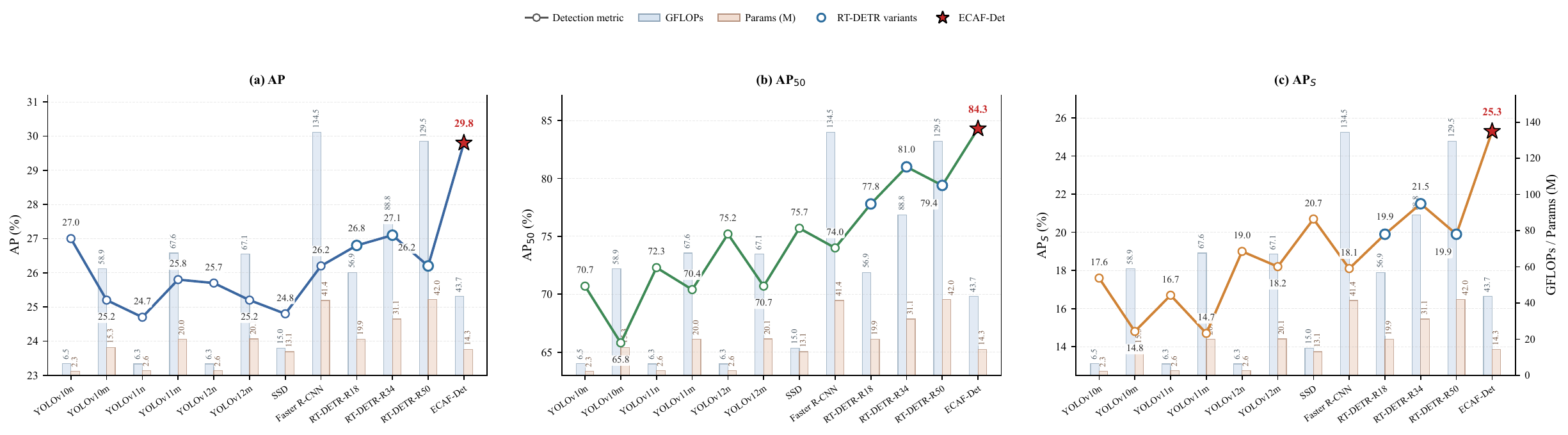}
	\caption{
		Performance and computational-complexity comparison of ECAF-Det and
		representative detectors on the IIG dataset. Panels (a)--(c) show AP,
		AP$_{50}$, and AP$_{S}$, respectively. In each panel, the line represents
		detection performance, while the bars indicate GFLOPs and parameter counts.
		Numerical values are annotated for all compared methods to facilitate direct
		performance--complexity comparison.
	}
	\label{fig:1}
\end{figure}
To address these challenges, this study proposes an Edge-Aware and Content-Adaptive Feature Fusion Detector (ECAF-Det) for infrared gas leak detection in industrial monitoring scenarios. The proposed detector is designed to enhance weak plume features, preserve boundary-sensitive information, and adaptively fuse multi-scale representations while maintaining moderate computational cost. Instead of treating gas leakage as a standard rigid-object detection problem, ECAF-Det focuses on the visual characteristics of infrared gas plumes, including weak contrast, diffuse boundaries, semi-transparency, and large scale variation. {
	Figure~1 provides an initial comparison between ECAF-Det and several
	representative detectors on the IIG dataset in terms of AP, AP$_{50}$,
	AP$_{S}$, GFLOPs, and parameter count.
} {Within the RT-DETR detection framework, this study focuses on task-oriented
	feature representation and multi-scale feature aggregation for infrared gas
	plumes with weak contrast, diffuse structure, semi-transparency, and large
	scale variation.} The main contributions of this study are summarized as follows:

\begin{itemize}
	\item[\textcolor{black}{$\bullet$}] 
	A plume-oriented local--global feature enhancement block is developed for infrared gas plume representation. The local branch preserves fine-grained texture and boundary cues, while the global branch captures long-range contextual continuity of diffuse plume regions. This design improves the representation of faint and low-contrast gas regions in complex thermal backgrounds.
	
	\item[\textcolor{black}{$\bullet$}] 
	{
		A multi-scale edge perception module is introduced to provide
		boundary-sensitive auxiliary cues for semi-transparent gas plumes.
		Directional-gradient and Gabor-response cues are converted into hierarchical
		structural priors and incorporated into the backbone to strengthen the
		perceptibility and representation of weak plume regions. These priors are
		not explicitly supervised as plume-boundary maps; instead, their task
		relevance is determined through the downstream detection objective.
	}
	
	\item[\textcolor{black}{$\bullet$}] 
	{A content-adaptive sparse routing path aggregation network is designed for efficient multi-scale feature fusion. By estimating the importance of feature responses, the proposed routing mechanism assigns larger contributions to informative plume-related features while limiting the contribution of less informative responses during cross-scale aggregation.}

\end{itemize}

\section{Related Work}

\subsection{Conventional Gas Leak Detection and Infrared Imaging}
Gas leak detection methods can generally be divided into contact-based and non-contact methods~\citep{weng2024enhancing,li2024lr}. Contact-based methods mainly rely on chemical gas sensors, which detect target gases through direct interaction and can provide high sensitivity under controlled conditions~\citep{kang2023feature}. However, their performance is often affected by environmental factors such as temperature, humidity, airflow, and sensor aging~\citep{vergara2012chemical}. In addition, because most chemical sensors provide point-wise measurements, they have limited spatial coverage and usually cannot directly localize the leakage source without additional inspection or dense sensor deployment.

Non-contact methods infer gas leakage from physical signals such as optical, acoustic, or pressure responses~\citep{yao2024tsff,yao2025spatio,zhao2024review}. Acoustic-based approaches detect leakage-induced acoustic emissions and have been widely studied for pipeline monitoring. For example, ConvFormer~\citep{li2025optimized} combines signal denoising with deep classification to improve leak identification. Nevertheless, acoustic methods remain sensitive to background noise, machinery vibration, sensor placement, and domain shifts between operating conditions. Pressure-wave-based methods can provide fast response and accurate localization in well-defined pipeline systems~\citep{chen2024fbg}, but their performance may degrade because of wave attenuation, pressure fluctuations, pipeline complexity, and the need for multiple sensors or access points. These limitations make wide-area monitoring in large industrial facilities difficult.

Infrared (IR) imaging has become an important non-contact technique for industrial gas leak monitoring because it can provide spatially resolved plume information without direct contact with the leakage source. Long-wave and mid-wave infrared cameras can visualize gas plumes under suitable imaging conditions and can support day--night monitoring in hazardous or difficult-to-access environments~\citep{meribout2021gas}. In practice, however, IR gas leak inspection still often relies on manual visual interpretation, which is labor-intensive, operator-dependent, and prone to false alarms or missed detections. Low thermal contrast, adverse weather, complex industrial backgrounds, and semi-transparent gas plumes further reduce plume visibility and make reliable detection difficult~\citep{zimmerle2020detection}. These challenges motivate the integration of IR imaging with automated visual detection algorithms for industrial gas leak monitoring.

\subsection{Deep Learning for Infrared Gas Leak Detection}
Deep learning has been increasingly applied to infrared gas leak detection, with most existing methods built upon general object detection or video analysis frameworks. Conventional two-stage detectors, such as R-CNN~\citep{girshick2014rich}, Fast R-CNN~\citep{girshick2015fast}, and Faster R-CNN~\citep{ren2017fasterrcnn}, first generate region proposals and then perform classification and bounding-box refinement. These methods can achieve accurate localization, but their relatively high computational cost limits their suitability for real-time industrial monitoring. For infrared gas leakage scenes, TSFF-Net~\citep{yao2024tsff} combines temporal motion information and spatial appearance features to improve SF$_6$ leak detection in complex backgrounds. However, methods that rely strongly on temporal motion cues may be less effective for weak, slowly varying, or near-static plume appearances, and their sequential processing can increase deployment complexity. 

{From the perspective of engineering deployment, lightweight deep learning
	frameworks have also been investigated for gas-pipeline leak detection on
	low-power edge devices. For example, \citet{shi2026gasleakedge} developed an edge-oriented gas-leak detection
	framework and reported an average response time of 46.45 ms with a model
	parameter storage of 8.11 MB. This study highlights the importance of
	considering not only detection accuracy, but also model footprint, inference
	latency, and target hardware when assessing practical deployment feasibility.}

One-stage detectors, including YOLO and SSD~\citep{jiang2022review,liu2016ssd}, directly predict object categories and bounding boxes from dense feature maps, offering simpler architectures and faster inference. Several task-specific gas detection models have been developed on this basis. GasNet~\citep{wang2020machine} and VideoGasNet~\citep{wang2022videogasnet} have shown promising performance for methane leak detection and grading in video sequences. More recent methods, such as SRHS-Net~\citep{pan2025srhs} and DLFANet~\citep{jing2025lightweight}, introduce attention mechanisms, multi-scale receptive fields, and lightweight feature extraction to improve the detection of small vapor or gas targets under complex backgrounds. Deep feature learning and feature fusion have also been shown to be effective
for small-object detection, which is consistent with the need to preserve weak
and small plume features in infrared gas leak monitoring~\citep{tong2024smallobject}. These studies demonstrate the feasibility of deep learning for visual gas leak detection. Nevertheless, CNN-based methods may still struggle when gas plumes are faint, semi-transparent, weakly bounded, or partially blended with thermal clutter. In addition, fixed anchors, fixed receptive fields, and static multi-scale fusion strategies can limit their adaptability to plume targets with large variations in size, shape, and contrast.

Transformer-based detectors provide another direction for gas plume detection because self-attention can capture long-range contextual dependencies. DETR~\citep{carion2020end} formulates object detection as a set prediction problem and removes the need for hand-crafted anchors and proposal generation. Its variants, including Conditional DETR~\citep{meng2021conditional}, DN-DETR~\citep{li2022dn}, and RT-DETR~\citep{zhao2024detrs}, improve convergence speed, training stability, and inference efficiency through improved query design, denoising training, and hybrid encoder structures. These detectors are attractive for infrared gas monitoring because gas plumes often require contextual information beyond local texture cues. However, transformer-based detectors can still miss small or low-contrast plumes if weak plume responses are suppressed during backbone extraction or multi-scale feature aggregation. Therefore, task-oriented feature enhancement remains necessary for infrared gas leak detection.

{Frequency-domain feature modeling provides another approach to enlarging
	the effective interaction range and exploiting global contextual information.
	FNet replaces self-attention with parameter-free Fourier-domain token
	mixing~\citep{leethorp2022fnet}, while GFNet performs global spatial
	interaction through Fourier transformation, learnable frequency-domain
	filtering, and inverse transformation~\citep{rao2021global}. AFNO further
	introduces adaptive Fourier-domain token mixing with frequency-mode
	sparsification~\citep{guibas2022adaptive}. From the perspective of
	frequency-aware visual representation, FcaNet employs discrete cosine
	transform (DCT) components to construct multi-spectral channel
	attention~\citep{qin2021fcanet}. These studies demonstrate the effectiveness
	of frequency-domain transformations and spectral modulation for global or
	frequency-aware feature modeling. Motivated by this general line of research,
	our local--global plume block uses DCT-domain low-frequency modulation as a
	contextual modeling mechanism and couples it with local detail extraction
	to accommodate the diffuse and weakly contrasted characteristics of
	infrared gas plumes.}

Beyond detector architectures, edge-aware representation and multi-scale feature fusion are important for detecting faint and diffuse gas plumes. Classical edge operators, such as Sobel, Canny, and Laplacian~\citep{heath1998comparison,canny1986computational,wang2007laplacian}, can highlight intensity transitions and structural boundaries, but they are sensitive to noise and may fail when plume boundaries are weak, semi-transparent, or mixed with background thermal textures. Modern detectors commonly use feature pyramid structures, including FPN~\citep{lin2017feature}, PANet~\citep{liu2018path}, BiFPN~\citep{tan2020efficientdet}, and NAS-FPN~\citep{ghiasi2019fpn}, to combine low-level spatial details with high-level semantic information. However, most of these fusion strategies use fixed pathways or globally learned fusion weights, making them less responsive to spatially varying plume characteristics. For infrared gas leakage images, an effective fusion strategy should
adaptively emphasize weak plume-related features while regulating the
contribution of less informative responses from cluttered thermal
backgrounds. In related engineering AI studies on thermal infrared perception,
attention-based feature fusion has been used to improve object detection
performance under challenging imaging conditions, further supporting the need
for adaptive feature fusion in weak infrared plume detection~\citep{kowalski2025bispectral}.

{
	Overall, the reviewed studies indicate that weak infrared gas plume
	detection requires feature representations that preserve weak structural
	cues, exploit broader contextual information, and adapt cross-scale feature
	propagation. Existing approaches typically address only part of these
	requirements, while fixed or globally weighted feature-fusion strategies
	may remain insufficiently responsive to the diffuse structure, weak contrast,
	and large scale variation of infrared gas plumes. These gaps motivate the
	task-oriented feature representation and adaptive multi-scale aggregation
	developed in the following section.
}
\begin{figure}
	\centering
	\includegraphics[width=1\textwidth]{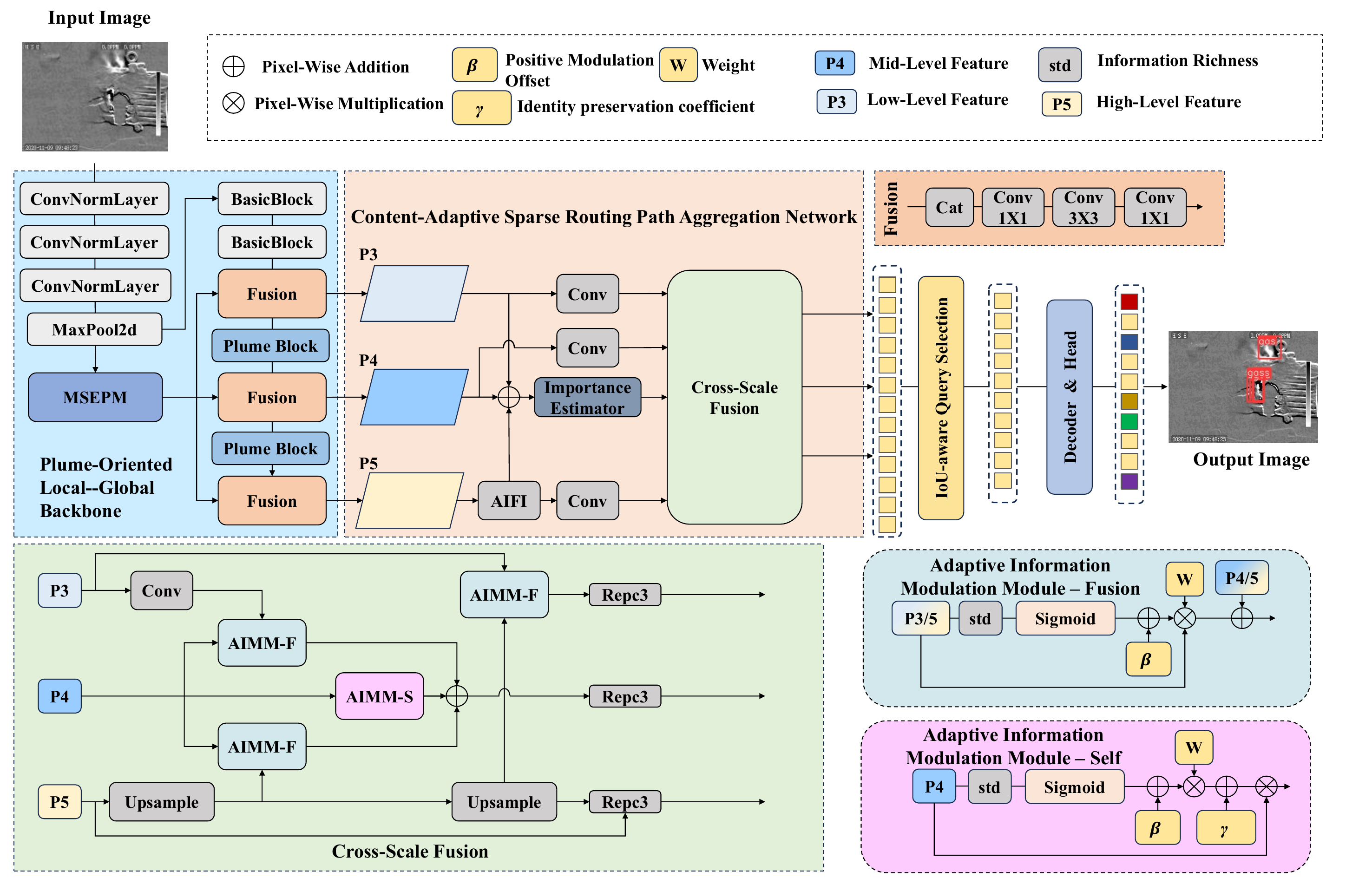}
	\caption{Overall architecture of the proposed Edge-Aware and Content-Adaptive Feature Fusion Detector (ECAF-Det) for infrared gas leak detection. 
		The framework consists of a multi-scale edge perception module (MSEPM), a plume-oriented local--global backbone, a content-adaptive sparse routing path aggregation network (CASR-PAN), and a decoder-based detection head. 
		MSEPM provides hierarchical boundary-sensitive structural priors to support
		plume representation, while CASR-PAN uses an importance estimator to generate content-dependent routing weights for adaptive multi-scale feature aggregation. 
		In AIMM-F and AIMM-S, $\beta$ denotes a fixed positive offset in the
		variation-dependent modulation term, while $\gamma$ denotes the fixed
		identity-preservation coefficient.}
	\label{fig:2}
\end{figure}
\section{Methodology}

\subsection{Overview of the Gas Leak Detection Framework}

{
	The proposed Edge-Aware and Content-Adaptive Feature Fusion Detector
	(ECAF-Det) is built upon the RT-DETR framework. It is designed for infrared
	gas leak detection in weak-plume and cluttered thermal scenes, as illustrated
	in Figure~\ref{fig:2}.
} Instead of treating gas leakage as a conventional rigid-object detection problem, ECAF-Det focuses on the visual characteristics of infrared gas plumes, including weak contrast, diffuse boundaries, semi-transparency, and large scale variation. {The query-selection and decoder-based detection mechanisms follow RT-DETR,
	while the principal modifications are introduced in the backbone feature
	representation and the multi-scale aggregation pathway.}

\begin{figure}
	\centering
	\includegraphics[width=1\textwidth]{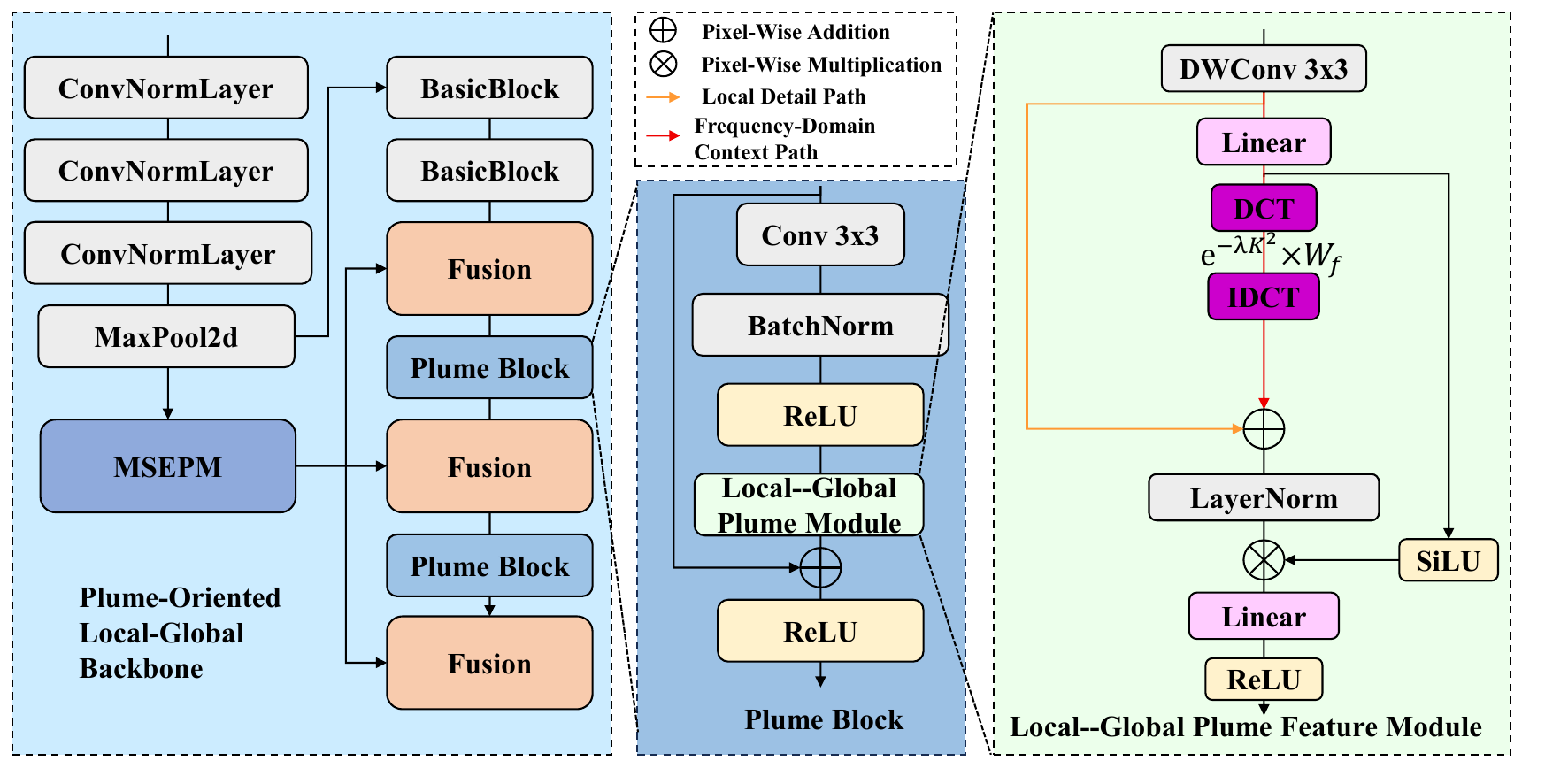}
	\caption{{
			Structure of the plume-oriented local--global backbone.
			MSEPM generates hierarchical boundary-sensitive structural priors,
			which are fused with the corresponding backbone features through
			scale-specific feature-fusion operations. The local--global plume
			blocks then combine a local branch for preserving fine-grained plume
			structures with a frequency-domain global branch for capturing broader
			contextual continuity.
	}}
	\label{fig:3}
\end{figure}

The framework contains three task-oriented components. First, the plume-oriented local--global backbone enhances weak plume representation by combining fine-grained local feature extraction with long-range contextual modeling. As shown in Figure~\ref{fig:3}, the local branch preserves texture details and boundary cues that are important for small and faint plumes, while the global branch captures contextual continuity across diffuse plume regions. {
	The local-detail and frequency-domain contextual branches are combined
	within the plume block, while the MSEPM-derived structural priors are
	fused with the corresponding backbone-stage features through
	scale-specific feature-fusion operations.
}	
{
	Second, the multi-scale edge perception module (MSEPM) generates hierarchical
	boundary-sensitive structural priors from directional-gradient and
	Gabor-response cues.
	These priors are introduced into different
	backbone stages as auxiliary structural cues rather than as explicitly
	supervised plume-edge predictions. Their contribution to plume
	representation is optimized jointly with the detector through the
	end-to-end detection objective.
}
Third, the content-adaptive sparse routing path aggregation network (CASR-PAN) performs adaptive multi-scale feature fusion before the decoder. Guided by an importance estimator (IE), CASR-PAN assigns content-dependent routing weights to different feature paths and selectively propagates informative plume-related features across scales. The adaptive information modulation module for fusion (AIMM-F) regulates cross-scale feature aggregation, while the adaptive information modulation module for self-enhancement (AIMM-S) refines features within the same scale. {Through this design, ECAF-Det enhances discriminative plume features, adaptively regulates cross-scale information propagation, and maintains moderate computational cost for practical infrared gas leak monitoring.}

\subsection{Local--Global Plume Block}
Infrared gas plumes usually exhibit two visual characteristics that are difficult to capture with standard convolutional blocks. On the one hand, small and faint plumes contain weak local texture and boundary cues that can be easily suppressed during feature extraction. On the other hand, gas regions often appear as diffuse and spatially continuous structures, requiring a larger contextual field to preserve plume integrity. {
	To address these two requirements, we design a local--global plume block
	that combines local detail preservation and long-range contextual aggregation.
}

Given an input feature map $X \in \mathbb{R}^{B \times C \times H \times W}$, the local branch first applies depthwise convolution to enhance spatially local plume variations:
\begin{equation}
	X_{\mathrm{local}} = \mathrm{DWConv}(X).
\end{equation}
The depthwise operation preserves channel-specific spatial responses and is used to capture fine-grained plume texture, weak boundary variations, and small-scale contrast changes. The output is then projected and split into a feature component and a modulation component:
\begin{equation}
	X_{\mathrm{proj}}, Z = \mathrm{split}\left(\mathrm{Linear}(X_{\mathrm{local}}), 2\right),
\end{equation}
where $Z$ provides a learnable gating signal for subsequent feature modulation.

To capture long-range contextual continuity, the global branch operates in the frequency domain. The projected feature $X_{\mathrm{proj}}$ is transformed by the discrete cosine transform (DCT), modulated by a learnable frequency weighting function, and then transformed back to the spatial domain:
\begin{equation}
	X_{\mathrm{dct}} = \mathrm{DCT}(X_{\mathrm{proj}}),
\end{equation}
\begin{equation}
	X_{\mathrm{dct}}^{\mathrm{mod}} = X_{\mathrm{dct}} \odot \exp(-\lambda K^2) \odot W_f,
\end{equation}
\begin{equation}
	X_{\mathrm{global}} = \mathrm{IDCT}\left(X_{\mathrm{dct}}^{\mathrm{mod}}\right),
\end{equation}
{where $K^2=\omega_x^2+\omega_y^2$,
	$\omega_x=\pi k_x/W$, and $\omega_y=\pi k_y/H$.
	The coefficient $\lambda$ controls the attenuation strength, and $W_f$
	denotes the learnable channel-wise frequency weight. The exponential term
	$\exp(-\lambda K^2)$ forms a Gaussian-shaped low-pass weighting in the
	DCT domain, progressively attenuating higher-frequency components as the
	spatial frequency increases. In the proposed plume block, this operation
	provides a low-frequency contextual bias that is suitable for spatially
	smooth and diffuse plume responses. The resulting contextual representation is subsequently combined with
	the local-detail branch, allowing the block to jointly retain weak local
	structures and broader plume context.}

{Finally, the local-detail and frequency-domain contextual features are
	combined and projected back to the original feature dimension:
	\begin{equation}
		Y' =
		\operatorname{OutLinear}
		\left(
		\operatorname{OutNorm}
		\left(
		X_{\mathrm{local}} + X_{\mathrm{global}}
		\right)
		\odot \operatorname{SiLU}(Z)
		\right),
	\end{equation}
	\begin{equation}
		Y = \operatorname{Act}(Y' + X_{\mathrm{res}}),
	\end{equation}
	where $X_{\mathrm{res}}$ denotes the residual input,
	$\operatorname{OutNorm}(\cdot)$ denotes normalization, and
	$\operatorname{Act}(\cdot)$ denotes the activation function.}

The proposed local--global plume block is designed as a task-oriented feature enhancement module rather than a numerical solver for gas transport equations. The local branch focuses on fine plume details and boundary cues, while the frequency-domain global branch improves long-range contextual continuity. {Through the complementary local-detail and frequency-domain contextual
	branches, the block enhances faint and diffuse plume representations
	under complex thermal backgrounds.}

\subsection{Multi-Scale Edge Perception Module}

Infrared gas plumes are often semi-transparent and weakly contrasted against the thermal background.
Their boundaries are usually diffuse rather than sharp, and the intensity transition between the plume
and the surrounding scene can be very small. As a result, standard convolutional features may
underrepresent weak plume boundaries or confuse background thermal structures with gas-related
edges. To provide auxiliary boundary-sensitive cues for weak plume detection, we introduce a
multi-scale edge perception module (MSEPM).
\begin{figure}
	\centering
	\includegraphics[width=0.8\textwidth]{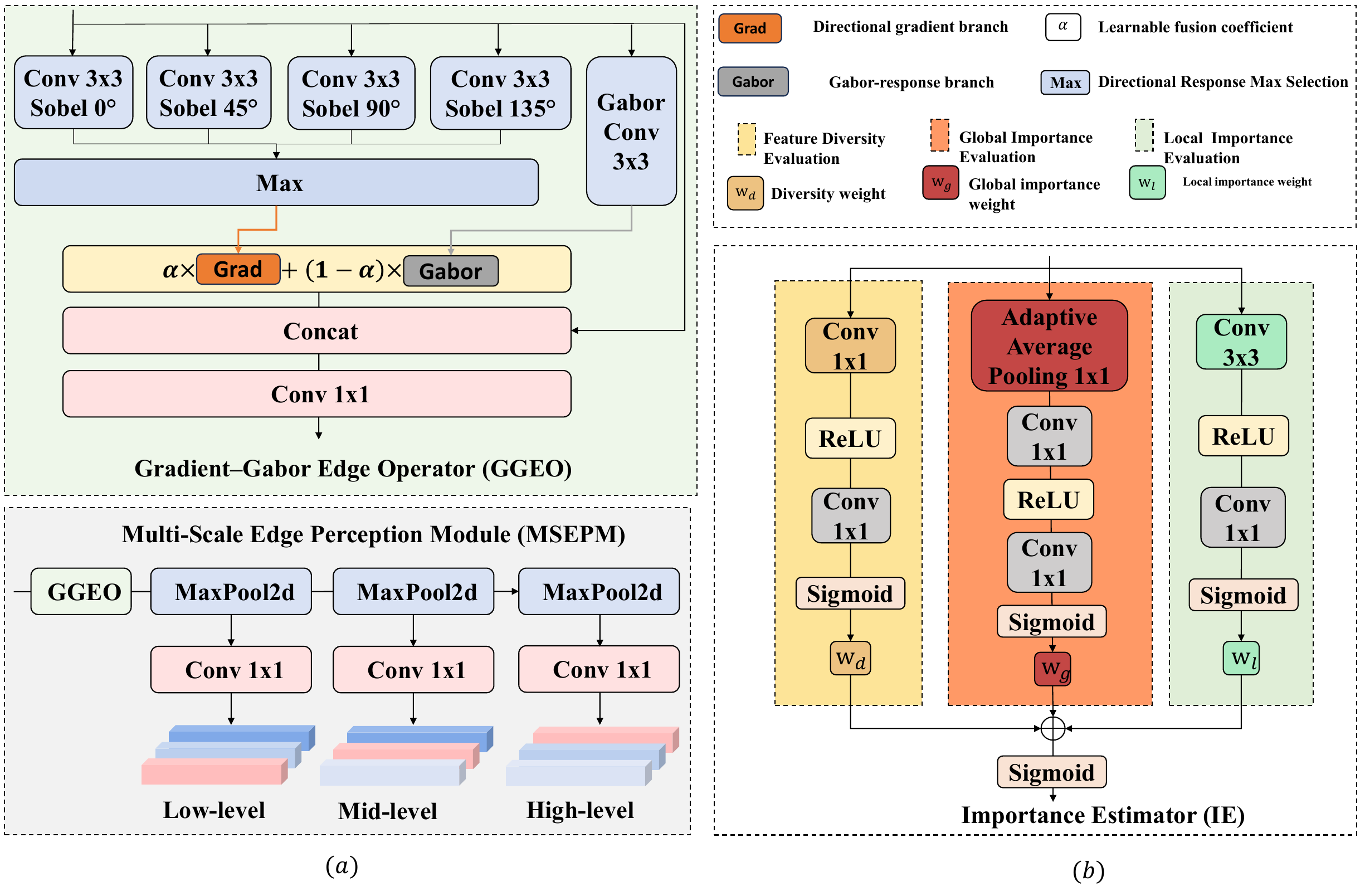}
	\caption{
		Architectures of the proposed structural-prior generation and
		content-adaptive routing components.
		(a) Multi-scale edge perception module (MSEPM) and gradient--Gabor
		edge operator (GGEO). GGEO combines four fixed $3\times3$ Sobel-like
		directional-gradient kernels with a depthwise $3\times3$ Gabor-response
		branch to generate an initial structural representation $E_0$.
		MSEPM then constructs hierarchical structural priors through progressive
		$2\times2$ max-pooling and $1\times1$ projection.
		(b) Importance estimator (IE) in CASR-PAN. The estimator integrates
		global importance, local importance, and feature-diversity cues to
		generate content-dependent importance maps for adaptive multi-scale
		feature routing.
	}
	\label{fig:4}
\end{figure}		
The MSEPM first employs a gradient–Gabor edge operator (GGEO) to combine directional-gradient and Gabor-filter responses into an initial structural representation. Progressive downsampling and scale-specific projection then transform this representation into hierarchical structural priors for the backbone. Given an input feature map $(X \in \mathbb{R}^{B \times C \times H \times W})$, the directional-gradient branch applies four fixed $(3 \times 3)$ Sobel-like kernels with orientations of $(0^\circ)$, $(45^\circ)$, $(90^\circ)$, and $(135^\circ)$. The aggregated response is defined as follows:
\begin{equation}
	G(x,y) =
	\max_{\theta \in \{0^{\circ},45^{\circ},90^{\circ},135^{\circ}\}}
	\left|K_{\theta} * X(x,y)\right|,
\end{equation}
where $*$ denotes depthwise convolution.		
{
	In the implementation, all directional kernels are fixed $3\times3$
	Sobel-like filters. Their numerical weights are}
\begin{equation}
	\label{eq:directional_kernels}
	\begin{aligned}
		K_{0^\circ}
		&=
		\begin{bmatrix}
			1 & 0 & -1\\
			2 & 0 & -2\\
			1 & 0 & -1
		\end{bmatrix},
		&
		K_{45^\circ}
		&=
		\begin{bmatrix}
			0 & 1 & 2\\
			-1 & 0 & 1\\
			-2 & -1 & 0
		\end{bmatrix},
		\\[4pt]
		K_{90^\circ}
		&=
		\begin{bmatrix}
			1 & 2 & 1\\
			0 & 0 & 0\\
			-1 & -2 & -1
		\end{bmatrix},
		&
		K_{135^\circ}
		&=
		\begin{bmatrix}
			2 & 1 & 0\\
			1 & 0 & -1\\
			0 & -1 & -2
		\end{bmatrix}.
	\end{aligned}
\end{equation}
{	The four kernels are applied independently to each input channel using
	depthwise convolution with stride 1 and padding 1. Their weights are fixed
	throughout training. The absolute response is computed for each orientation,
	and the maximum response across the four directions is retained to form
	$G(x,y)$.	
	To complement the directional-gradient response, the second branch employs
	a depthwise Gabor convolution. The Gabor kernel is initialized according to	}
\begin{equation}
	K_G(x,y)=
	\exp\left(
	-\frac{x_\theta^2+\gamma_G^2y_\theta^2}
	{2\sigma_G^2}
	\right)
	\cos\left(
	\frac{2\pi x_\theta}{\lambda_G}
	\right),
	\label{eq:gabor_kernel}
\end{equation}	
where
\begin{equation}
	x_\theta=x\cos\theta+y\sin\theta,
	\qquad
	y_\theta=-x\sin\theta+y\cos\theta .
	\label{eq:gabor_coordinates}
\end{equation}	
{	In our implementation, a single $3\times3$ Gabor kernel is initialized
	with orientation $\theta=0^\circ$, Gaussian standard deviation
	$\sigma_G=2.0$, wavelength $\lambda_G=4.0$, and aspect ratio
	$\gamma_G=0.5$. The cosine phase offset is zero. Before being assigned
	to the depthwise convolution, the initialized kernel is normalized by its
	absolute sum,
	$K_G \leftarrow K_G / \sum_{x,y}|K_G(x,y)|$.
	The same initialization is replicated across all input channels. The
	depthwise Gabor convolution is applied with stride 1 and padding 1.
	Unlike the fixed directional-gradient kernels, the Gabor convolution
	remains learnable during end-to-end training.		}		
The magnitude of the Gabor response is computed as
\begin{equation}
	R_G(x,y)=\left|K_G * X(x,y)\right|.
\end{equation}		
{
	The directional-gradient and Gabor responses are adaptively combined using
	a learnable coefficient. Specifically, the effective fusion coefficient is
	defined as $\alpha=\sigma(a)$, where $a$ is an unconstrained learnable
	scalar and $\sigma(\cdot)$ denotes the sigmoid function. The initial value
	of $a$ is set to 0.7, corresponding to an initial effective fusion
	coefficient of $\alpha=\sigma(0.7)\approx0.668$.
	The fused structural response is then concatenated with the original input
	feature and projected through a $1\times1$ convolution:	}
\begin{equation}
	E_0 =
	\mathrm{Conv}_{1\times1}
	\left(
	\mathrm{Concat}
	\left[
	\alpha G + (1-\alpha)R_G,\,
	X
	\right]
	\right).
\end{equation}	
Based on the initial GGEO output $E_0$, MSEPM constructs hierarchical
structural priors through progressive downsampling:
\begin{equation}
	E_i = \mathrm{Pool}(E_{i-1}),
	\qquad i=1,2,\dots,N,
\end{equation}
where $\mathrm{Pool}(\cdot)$ denotes a $2\times2$ max-pooling operation
with stride 2. In the implementation used in this study, $N=3$, producing
three progressively downsampled representations. Each downsampled feature
is subsequently projected by a $1\times1$ convolution to match the channel
dimension required by the corresponding backbone stage:
\begin{equation}
	\hat{E}_i =
	\mathrm{Conv}_{1\times1}^{(i)}(E_i),
	\qquad i=1,2,\dots,N.
\end{equation}		
{
	Therefore, the MSEPM outputs
	$\{\hat{E}_1,\hat{E}_2,\hat{E}_3\}$ rather than directly using $E_0$ as
	a backbone-stage output. These progressively downsampled and projected
	representations provide hierarchical boundary-sensitive structural priors
	for the corresponding backbone stages.	}		
{
	The structural responses generated by GGEO are not directly supervised by
	pixel-level plume-boundary annotations and therefore do not constitute
	semantic plume-edge predictions. Directional-gradient and Gabor filters may
	respond to both plume boundaries and background thermal structures. MSEPM
	consequently treats these responses as boundary-sensitive structural priors.
	Their task relevance is learned indirectly through the end-to-end detection
	objective, which propagates gradients through the feature-fusion operations,
	the learnable scalar $a$, the learnable Gabor convolution, the scale-specific
	projection layers, and downstream backbone parameters.	}

\subsection{Content-Adaptive Sparse Routing Path Aggregation Network}
Feature pyramid networks, such as FPN and PANet, commonly use fixed top-down and bottom-up pathways to fuse multi-scale features. Although these structures are effective for general object detection, their fusion paths are usually predefined and may not adapt well to infrared gas plumes with weak contrast, diffuse boundaries, and large scale variations. BiFPN introduces learnable fusion weights, and NAS-FPN searches for feature fusion structures, but their connections are still largely determined at the network level rather than dynamically adjusted according to the content of each feature map.

{Building on path-aggregation-based multi-scale fusion, we introduce
	content-adaptive routing for infrared gas plume features. The resulting
	content-adaptive sparse routing path aggregation network (CASR-PAN) uses
	an importance estimator (IE) to generate input-dependent routing weights
	for individual propagation paths according to the current feature content.
	Unlike fixed cross-scale aggregation or globally learned fusion coefficients,
	these routing weights vary with the input feature responses and dynamically
	regulate the contribution of propagated information during multi-scale
	aggregation. Larger routing weights assign greater contributions to
	informative cross-scale features, whereas smaller routing weights limit the
	contribution of less relevant features.} The routed features are further refined by RepC3 modules before being fed into the RT-DETR decoder. The structure of the importance estimator is shown in Figure~\ref{fig:4}(b).

Given an input feature map 
$\mathbf{X} \in \mathbb{R}^{B \times C \times H \times W}$, 
the importance estimator computes an importance map 
$\mathbf{I} \in \mathbb{R}^{B \times C \times H \times W}$ as
\begin{equation}
	\mathbf{I} =
	\sigma \left(
	w_g \tilde{\mathbf{G}}
	+ w_l \mathbf{L}
	+ w_d \mathbf{D}
	\right),
\end{equation}
where $\sigma(\cdot)$ denotes the sigmoid function. The coefficients $w_g$, $w_l$, and $w_d$ are learnable fusion weights normalized by a softmax function. The three terms $\tilde{\mathbf{G}}$, $\mathbf{L}$, and $\mathbf{D}$ represent global importance, local importance, and feature diversity, respectively.

The global importance term $\tilde{\mathbf{G}}$ is obtained by applying global average pooling to $\mathbf{X}$, followed by two $1\times1$ convolutional layers with a ReLU activation between them. The output is then activated by a sigmoid function and upsampled to the original spatial resolution. This branch captures channel-level responses that are globally informative for the current feature map.
The local importance term $\mathbf{L}$ is computed using a $3\times3$ convolution followed by ReLU activation, a $1\times1$ convolution, and a sigmoid function. Because the $3\times3$ convolution operates on local neighborhoods, this branch captures spatially localized plume cues, such as weak boundaries, small texture variations, and local contrast changes.
The diversity term $\mathbf{D}$ is designed to highlight channels that contain complementary or non-redundant information. It is computed using two $1\times1$ convolutional layers with a ReLU activation between them and is further modulated by the channel-wise standard deviation of $\mathbf{X}$. This term helps the estimator retain feature channels with higher response variation, which can be useful for distinguishing plume-related patterns from redundant background responses.
The aggregated importance map $\mathbf{I}$ is then transformed into four routing weights by a $1\times1$ convolution:
\begin{equation}
	\left[
	\mathbf{W}_1,
	\mathbf{W}_2,
	\mathbf{W}_3,
	\mathbf{W}_4
	\right]
	=
	\sigma
	\left(
	\mathrm{Conv}_{1\times1}(\mathbf{I})
	\right),
\end{equation}
where each $\mathbf{W}_i$ corresponds to one feature propagation path in CASR-PAN. These routing weights are used to regulate cross-scale fusion and self-enhancement paths.
Specifically, the four routing weights correspond to four explicit
feature-propagation paths in CASR-PAN. Let \(P_3\), \(P_4\), and
\(P_5\) denote the low-level, middle-level, and high-level pyramid
features, respectively. The routing weights can be written as
\begin{equation}
	[W_1,W_2,W_3,W_4]
	=
	[W_{5\rightarrow4},W_{5\rightarrow3},W_{3\rightarrow4},W_{4\rightarrow4}],
\end{equation}
{
	Here, \(W_{5\rightarrow4}\) controls the deep-to-middle semantic propagation
	from \(P_5\) to \(P_4\). The weight \(W_{5\rightarrow3}\) controls the
	deep-to-shallow semantic propagation from \(P_5\) to \(P_3\).
	\(W_{3\rightarrow4}\) controls the shallow-to-middle detail compensation
	from \(P_3\) to \(P_4\), while \(W_{4\rightarrow4}\) controls the
	self-enhancement of the middle-level feature \(P_4\).
}
Thus, the four weights are not global scalar coefficients shared by
all fusion operations, but content-dependent modulation maps assigned
to specific routing paths. This design allows CASR-PAN to selectively
propagate high-level semantic cues, preserve low-level spatial details,
and enhance middle-level plume representations according to the image content.
The term ``sparse'' in CASR-PAN refers to the restricted path-connectivity
pattern rather than hard activation sparsity. Specifically, only three
cross-scale propagation paths and one middle-level self-enhancement path are
explicitly constructed, rather than introducing dense pairwise connections
among all pyramid levels. The routing coefficients on these selected paths
remain continuous content-dependent weights; no hard binary pruning or
top-$k$ path selection is performed.

For cross-scale fusion, the adaptive information modulation module for fusion (AIMM-F) takes two aligned feature maps 
$\mathbf{F}_1, \mathbf{F}_2 \in \mathbb{R}^{B \times C \times H \times W}$ 
as input. The output is computed as
\begin{equation}
	\mathbf{Y}
	=
	\mathbf{F}_1
	+
	\mathbf{F}_2
	\odot
	\mathbf{W}
	\odot
	\left(
	\beta + \sigma(\mathrm{Std}(\mathbf{F}_2))
	\right),
\end{equation}
{where $\odot$ denotes element-wise multiplication, $\mathbf{W}$ is the routing weight generated by the importance estimator, and $\mathrm{Std}(\cdot)$ denotes the channel-wise feature variation. The coefficient $\beta$ is a fixed positive offset in the variation-dependent modulation term. In this work, $\beta$ is set to $0.5$, providing a positive offset to the feature-variation term during cross-scale fusion.}	
For self-enhancement, the adaptive information modulation module for self-enhancement (AIMM-S) refines a single feature map $\mathbf{F}$ as
\begin{equation}
	\mathbf{Y}
	=
	\mathbf{F}
	\odot
	\left[
	\gamma
	+
	\mathbf{W}
	\odot
	\left(
	\beta + \sigma(\mathrm{Std}(\mathbf{F}))
	\right)
	\right],
	\label{eq:aimm_s}
\end{equation}
{where $\gamma$ is a fixed identity-preservation coefficient. In this work, $\gamma$ is set to $1.0$ so that the original feature response is preserved while the routing weight provides content-dependent self-enhancement.
	Because $\gamma=1.0$, $\mathbf{W}\in(0,1)$, and $\beta+\sigma(\mathrm{Std}(\mathbf{F}))>0$, the multiplicative factor 
	$\gamma+\mathbf{W}\odot[\beta+\sigma(\mathrm{Std}(\mathbf{F}))]$ 
	in Eq.~\eqref{eq:aimm_s}
	is element-wise greater than $1$. Therefore, AIMM-S does not provide direct multiplicative attenuation relative to the identity mapping. Instead, it preserves the original middle-level representation and applies content-dependent enhancement, with regions assigned larger routing weights receiving stronger modulation. {
		The reduction of less relevant cross-scale information in CASR-PAN is
		achieved through contribution regulation in AIMM-F. Smaller routing weights
		reduce the contribution of incoming features during aggregation. This
		mechanism does not attenuate the existing feature response in AIMM-S.
	}
	It should be noted that $\beta$ and $\gamma$ are fixed stabilization coefficients rather than adaptive routing parameters. The adaptive behavior of AIMM-F and AIMM-S is determined by the content-dependent routing weight $\mathbf{W}$ generated by the importance estimator. Accordingly, AIMM-F regulates information exchange between different scales by controlling the contribution of incoming features, whereas AIMM-S preserves the identity response and selectively strengthens discriminative representations within the same scale. {The sensitivity of these two fixed coefficients is evaluated in Section~4.5.9 using a controlled CASR-PAN-only setting.}}

CASR-PAN constructs three cross-scale routing paths and one self-enhancement path. Specifically, high-level features are propagated to middle- and low-level features to provide semantic context, while low-level features are propagated to middle-level features to preserve spatial details. The self-enhancement path further refines the middle-level representation. After routing and modulation, features at the same scale are aggregated and refined by RepC3 modules, producing the final multi-scale features for the decoder.

{Through this design, CASR-PAN performs content-adaptive multi-scale fusion for infrared gas plume detection. In AIMM-F, larger routing weights increase the contribution of informative incoming features during cross-scale aggregation, whereas smaller routing weights reduce the contribution of less relevant incoming features. In AIMM-S, the routing signal is used for selective self-enhancement on top of the identity-preserved representation rather than for direct attenuation. This complementary design allows CASR-PAN to adaptively regulate cross-scale information exchange while strengthening discriminative middle-level plume representations.}

\subsection{Training Objective}

The proposed ECAF-Det follows the RT-DETR detection training paradigm.
The overall training objective consists of classification loss, bounding-box
regression loss, generalized IoU loss, and auxiliary decoder losses:
\begin{equation}
	\mathcal{L}
	=
	\lambda_{\mathrm{cls}}\mathcal{L}_{\mathrm{cls}}
	+
	\lambda_{\mathrm{L1}}\mathcal{L}_{\mathrm{L1}}
	+
	\lambda_{\mathrm{giou}}\mathcal{L}_{\mathrm{giou}}
	+
	\lambda_{\mathrm{aux}}\mathcal{L}_{\mathrm{aux}} .
\end{equation}
Here, \(\mathcal{L}_{\mathrm{cls}}\) denotes the classification loss,
\(\mathcal{L}_{\mathrm{L1}}\) measures the bounding-box regression error,
\(\mathcal{L}_{\mathrm{giou}}\) denotes the generalized IoU loss, and
\(\mathcal{L}_{\mathrm{aux}}\) represents the auxiliary losses from
intermediate decoder layers. Hungarian matching is used to assign predicted
queries to ground-truth boxes during training. {
	No additional edge-, routing-, or mask-supervision loss is introduced for
	MSEPM or CASR-PAN. All trainable components of the proposed modules are
	optimized end-to-end through the detection objective.
}

\section{Experiments and Analysis}

\subsection{Datasets}
{Experiments were conducted on two infrared gas leakage datasets: IIG and
	LangGas. The IIG dataset, introduced by \citet{yu2024lightweight}, contains
	5,569 thermal gas images with 11,186 annotated gas leakage instances collected
	using a VF 330-VOC series handheld thermal infrared imaging device. According
	to the original dataset description, the images were acquired from five
	different industrial scenarios. All IIG images used in this study were taken
	directly from the released dataset; no synthetically generated IIG samples
	were introduced in either the training or test set.}

{
	Following the adopted 4:1 dataset split, the IIG dataset was divided into
	a training set and a test set. The training set contains 4,453 images with
	9,289 gas leakage instances, while the test set contains 1,116 images with
	1,897 instances. All quantitative results produced in this study on the IIG dataset were
	evaluated using this test set.}

The LangGas dataset~\citep{guo2025langgas} was further used to evaluate the performance of the proposed method on an additional infrared gas plume dataset. 
LangGas is a synthetic infrared gas leakage dataset generated by compositing physically realistic gas plumes and interfering foreground objects onto diverse background scenes. 
{Although LangGas provides pixel-level segmentation annotations and video
	sequences, this study focuses on object detection. Therefore, we manually
	annotated bounding boxes for the gas plume instances to construct the
	detection annotations used in this study. The original pixel-level masks were
	not converted into bounding boxes for the detection experiments. The resulting annotations were divided into 1,289 training images and 323
	test images.}

\subsection{Implementation Details}

All experiments were conducted under the same hardware and software environment to ensure fair comparison. 
The experiments were performed on a workstation equipped with an AMD Ryzen 7 9700X CPU and an NVIDIA GeForce RTX 5070 Ti GPU with 16 GB of memory. 
The software environment included Python 3.10.16, PyTorch 2.5.1, and CUDA 12.4.
{
	The inference latency and throughput values reported in Table~\ref{tab:industrial_alarm_metrics} were measured on the NVIDIA GeForce RTX 5070 Ti under the above experimental environment with a batch size of 4. The reported inference latency corresponds to the average model-forward time per image, whereas the reported throughput is calculated from the total per-image processing time including preprocessing, model inference, and postprocessing. Therefore, these measurements characterize inference efficiency on the evaluated workstation GPU platform and should not be interpreted as direct evidence of real-time performance on low-power edge devices or CPU-only platforms.
}

All input images were resized to $640 \times 640$. 
The models were trained for 200 epochs with a batch size of 4. 
AdamW was used as the optimizer with an initial learning rate of $1\times10^{-4}$, $\beta_1=0.9$, and $\beta_2=0.999$. 
Unless otherwise specified, the same training schedule and input
resolution were used for all comparison methods reproduced in this study.

Model performance was evaluated following the COCO evaluation protocol~\citep{lin2014microsoft}. 
The reported detection metrics include AP, AP$_{50}$, AP$_{75}$, AP$_S$, AP$_M$, and AP$_L$. 
Computational complexity was measured using floating-point operations in GFLOPs, and model size was measured by the number of parameters in millions. 
RT-DETR-R18 was used as the main baseline because ECAF-Det is developed upon the RT-DETR detection framework. {To control for the effect of backbone capacity, RT-DETR-R34 and RT-DETR-R50 were additionally trained and evaluated under the same experimental protocol used for RT-DETR-R18.}

\subsection{Comparison with Representative Detectors}
To evaluate the effectiveness of the proposed ECAF-Det, we compared it with several representative object detectors, including YOLOv10~\citep{wang2024yolov10}, YOLOv11~\citep{khanam2024yolov11}, YOLOv12~\citep{tian2025yolov12}, SSD~\citep{liu2016ssd}, Faster R-CNN~\citep{ren2017fasterrcnn} and RT-DETR~\citep{zhao2024detrs}. 
The general-purpose comparison models were trained and evaluated under
the same input resolution and dataset splits used in this study. 
The quantitative results on the IIG and LangGas datasets are summarized
in Tables~\ref{tab:comparison_mainstream} and
\ref{tab:comparison_mainstream_LangGas}, respectively.
For the IIG dataset, we additionally include the previously published
result of GasViT~\citep{yu2024lightweight}, a task-specific infrared gas
detection method developed on the same dataset. The literature-reported
result is marked with $^{\dagger}$ in
Table~\ref{tab:comparison_mainstream} and was not reproduced under our
experimental environment. Because the original publication does not report
the complete COCO metric set, unavailable metrics are denoted by ``--''. 
\begin{table}[htbp]
	\centering
	\caption{Detection performance comparison of ECAF-Det and representative detectors on the IIG dataset.}
	\label{tab:comparison_mainstream}
	\scriptsize
	\resizebox{\columnwidth}{!}{
		\begin{tabular}{lcccccccc}
			\toprule
			Model & AP (\%) & AP$_{50}$ (\%) & AP$_{75}$ (\%) & AP$_{S}$ (\%) & AP$_{M}$ (\%) & AP$_{L}$ (\%) & GFLOPs (G) & Params (M) \\
			\midrule
			YOLOv10n~\citep{wang2024yolov10}    &  27.0  &  70.7  &  10.6  &  17.6  &  \textbf{34.6}  &  38.4  &  6.5  &  2.265  \\
			YOLOv10m    &  25.2  &  65.8  &  \textbf{11.2}  &  14.8  &  34.2  &  23.6  &  58.9  &  15.313  \\
			YOLOv11n~\citep{khanam2024yolov11}    &  24.7  &  72.3  &  6.7  &  16.7  &  31.1  &  \textbf{42.8}  &  \textbf{6.3}  &  2.582 \\
			YOLOv11m    &  25.8  &  70.4  &  10.4  &  14.7  &  34.5  &  31.5  &  67.6  &  20.030  \\
			YOLOv12n~\citep{tian2025yolov12}    &  25.7  &  75.2  &  7.2  &  19.0  &  30.9  &  29.2  &  \textbf{6.3} &  \textbf{2.555} \\
			YOLOv12m    &  25.2  &  70.7  &  9.8  &  18.2  &  33.3  &  12.7  &  67.1  &  20.105 \\
			SSD~\citep{liu2016ssd}      &  24.8  &  75.7  &  6.5  &  20.7  &  29.0  &  23.3  &  15.03  &  13.06  \\
			Faster R-CNN~\citep{ren2017fasterrcnn}     &  26.2  &  74.0  &  9.6  &  18.1  &  33.7  &  24.0  &  134.491  &  41.352  \\			
			GasViT~\citep{yu2024lightweight}$^{\dagger}$
			& -- & 82.7 & -- & -- & -- & -- & -- & -- \\

			RT-DETR-R18~\citep{zhao2024detrs}      &  26.8  &  77.8  &  8.7  &  19.9  &  32.0  &  37.1  &  56.9  &  19.873  \\
			{RT-DETR-R34~\citep{zhao2024detrs}}      
			& {27.1}  
			& {81.0}  
			& {7.7}  
			& {21.5}  
			& {31.6}  
			& {36.4}  
			& {88.8}  
			& {31.106} \\
			
			{RT-DETR-R50~\citep{zhao2024detrs}}      
			& {26.2}  
			& {79.4}  
			& {7.2}  
			& {19.9}  
			& {31.2}  
			& {37.6}  
			& {129.5}  
			& {41.956} \\
			
			ECAF-Det     &\textbf{29.8}  &\textbf{84.3}  & 8.5 & \textbf{25.3} & 32.5 & 42.6 &43.7  &14.277  \\
			\bottomrule
		\end{tabular}
	}\vspace{1mm}
	\parbox{\columnwidth}{\footnotesize
		$^{\dagger}$ The result is directly reported from the corresponding
		original publication and was not reproduced in this study. ``--''
		indicates that the corresponding metric was not reported in the
		original publication.
	}
\end{table}

\begin{table}[htbp]
	\centering
	\caption{Detection performance comparison of ECAF-Det and representative detectors on the LangGas dataset.}
	\label{tab:comparison_mainstream_LangGas}
	\scriptsize
	\resizebox{\columnwidth}{!}{
		\begin{tabular}{lcccccccc}
			\toprule
			Model & AP (\%) & AP$_{50}$ (\%) & AP$_{75}$ (\%) & AP$_{S}$ (\%) & AP$_{M}$ (\%) & AP$_{L}$ (\%) & GFLOPs (G) & Params (M) \\
			\midrule
			YOLOv10n~\citep{wang2024yolov10}    &  32.3  &  61.7  &  30.2  &  0.1  & 11.4  &  39.8  &  6.5  &  2.265  \\
			YOLOv10m    &  28.8  &  56.9  &  26.8  &  0.8  &  14.1  &  34.9  &  58.9  &  15.313  \\
			YOLOv11n~\citep{khanam2024yolov11}    &  33.5  &  65.1  &  30.1  &  1.7  &  14.7  &  40.5  &  \textbf{6.3}  &  2.582 \\
			YOLOv11m    &  32.6  &  65.1  &  28.1  &  1.6  &  14.5  &  39.8  &  67.6  &  20.030  \\
			YOLOv12n~\citep{tian2025yolov12}    &  34.4  &  65.7  &  31.0  &  1.1  &  15.3  & 41.3  &  \textbf{6.3}  &  \textbf{2.555} \\
			YOLOv12m    &  32.8  &  63.6 &  28.7  &  1.7  &  13.8  &  39.7  &  67.1  &  20.105 \\
			SSD~\citep{liu2016ssd}       &  30.7  &  64.8  &  25.3  &  1.9  &  11.2  &  37.6  &  15.03  &  13.06  \\
			Faster R-CNN~\citep{ren2017fasterrcnn}     &  30.3  &  62.1  &  24.0  &  1.1  &  10.7  &  38.1  &  134.491  &  41.352  \\			
			RT-DETR-R18~\citep{zhao2024detrs}      & 31.4   & 63.6  & 28.4   &2.6   & 12.0   & 38.6   &  56.9  &  19.873  \\
			{RT-DETR-R34~\citep{zhao2024detrs}}
			& {35.0}
			& {66.5}
			& {\textbf{33.4}}
			& {\textbf{5.6}}
			& {14.1}
			& {42.6}
			& {88.8}
			& {31.106} \\
			
			{RT-DETR-R50~\citep{zhao2024detrs}}
			& {34.9}
			& {67.1}
			& {31.1}
			& {2.9}
			& {14.2}
			& {42.6}
			& {129.5}
			& {41.956} \\
			
			ECAF-Det
			& \textbf{36.3}
			& \textbf{68.5}
			& {31.2}
			& {3.7}
			& \textbf{15.4}
			& \textbf{43.9}
			& 43.7
			& 14.277 \\
			\bottomrule
		\end{tabular}
	}
\end{table}
On the IIG dataset, ECAF-Det achieves the highest AP of 29.8\% and AP$_{50}$ of 84.3\%, improving the RT-DETR-R18 baseline by 3.0 and 6.5 percentage points, respectively. 
The improvement is particularly evident for small gas plumes, where ECAF-Det obtains an AP$_S$ of 25.3\%, exceeding RT-DETR-R18 by 5.4 percentage points. 
These results indicate that the proposed edge-aware representation and content-adaptive multi-scale fusion are effective for detecting faint and small plume targets.
{
	To further examine whether the performance gain of ECAF-Det can be explained by increased backbone capacity, RT-DETR-R34 and RT-DETR-R50 were additionally evaluated under the same experimental protocol. On the IIG dataset, RT-DETR-R34 and RT-DETR-R50 achieve AP values of 27.1\% and 26.2\%, respectively, whereas ECAF-Det achieves 29.8\%. ECAF-Det also achieves higher AP$_{50}$ and AP$_S$ than both stronger RT-DETR variants. On LangGas, RT-DETR-R34 and RT-DETR-R50 improve the AP of the R18 baseline from 31.4\% to 35.0\% and 34.9\%, respectively, but both remain below the 36.3\% AP obtained by ECAF-Det. RT-DETR-R34 achieves higher AP$_{75}$ and AP$_S$ than ECAF-Det on LangGas, indicating that increased backbone capacity can benefit strict localization and small-scale plume detection in some cases. However, ECAF-Det requires only 14.277 M parameters and 43.7 GFLOPs, compared with 31.106 M parameters and 88.8 GFLOPs for RT-DETR-R34 and 41.956 M parameters and 129.5 GFLOPs for RT-DETR-R50. These results indicate that the overall performance improvement of ECAF-Det cannot be attributed simply to increasing the backbone capacity of RT-DETR.
}
\begin{table}[!t]
	\centering
	\caption{{
			Detection-oriented performance comparison between RT-DETR-R18 and
			ECAF-Det on the IIG test set. Model-inference latency and end-to-end
			throughput were measured on an NVIDIA GeForce RTX 5070 Ti GPU with a
			batch size of 4. End-to-end FPS includes preprocessing, model inference,
			and postprocessing.
	}}
	\label{tab:industrial_alarm_metrics}
	\resizebox{\textwidth}{!}{
		\begin{tabular}{lccccccc}
			\toprule
			\textbf{Model} 
			& \textbf{Precision (\%)} 
			& \textbf{Recall (\%)} 
			& \textbf{F1-score (\%)} 
			& \textbf{Missed detection rate (\%)} 
			& \textbf{Model inference latency (ms/image)}
			& \textbf{End-to-end FPS}
			&{\textbf{AP$_{50}$ (\%)}} \\
			\midrule
			RT-DETR-R18 
			& 86.97 
			& 70.60 
			& 77.94 
			& 29.40 
			& 12.53 
			& 75.60 
			& 77.8 \\
			ECAF-Det 
			& 86.31 
			& 79.44 
			& 82.73 
			& 20.56 
			& 15.46 
			& 61.81 
			& 84.3 \\
			\bottomrule
		\end{tabular}
	}
\end{table}

In terms of computational cost, ECAF-Det requires 43.7 GFLOPs and 14.277 M parameters, which is lower than RT-DETR-R18 and Faster R-CNN but higher than lightweight YOLO variants. 
This suggests that ECAF-Det provides a favorable accuracy--complexity trade-off among the evaluated detectors, although ultra-lightweight models such as YOLOv10n and YOLOv12n remain more efficient in terms of GFLOPs and parameter count. 

{For additional deployment context, the low-power gas-pipeline leak detection
	framework of \citet{shi2026gasleakedge} reports an average
	response time of 46.45 ms and a model parameter storage of 8.11 MB.
	However, they were obtained using a different sensing
	setting, implementation environment, and hardware platform from those used
	in the present infrared OGI detection experiments. 
	Under the controlled experimental setting of this study, ECAF-Det contains
	14.277 M parameters and requires 43.7 GFLOPs. {
		On the NVIDIA GeForce RTX 5070 Ti with a batch size of 4, ECAF-Det requires an average model-inference latency of 15.46 ms per image and achieves an end-to-end throughput of 61.81 FPS when preprocessing, model inference, and postprocessing are jointly considered.
	} More importantly,
	relative to the RT-DETR-R18 baseline evaluated under the same conditions,
	ECAF-Det reduces the parameter count from 19.873 M to 14.277 M and the
	computational cost from 56.9 GFLOPs to 43.7 GFLOPs, while improving
	weak- and small-plume detection.
	Direct evaluation and hardware-aware optimization on low-power
	industrial edge platforms are still required before its practical
	edge-device deployment can be established.}

{It is important to distinguish weak-plume detectability from strict
	high-IoU localization. On the IIG dataset, ECAF-Det improves AP from
	26.8\% to 29.8\%, AP$_{50}$ from 77.8\% to 84.3\%, and AP$_S$ from
	19.9\% to 25.3\% relative to RT-DETR-R18, whereas AP$_{75}$ changes
	from 8.7\% to 8.5\%. These results indicate that the principal improvement
	of ECAF-Det lies in detecting and recovering the approximate spatial extent
	of weak and small plume regions rather than in strict high-IoU bounding-box
	localization.}

{For weak infrared gas plumes, reliable plume detection is a prerequisite
	for subsequent precise localization. The substantial improvements in
	AP$_{50}$ and AP$_S$ suggest that the proposed boundary-sensitive structural
	guidance helps preserve plume responses that may otherwise be missed during
	feature extraction. However, the AP$_{75}$ result also shows that improved
	plume perceptibility does not automatically translate into more accurate
	high-IoU box alignment.}

{Because all compared methods are evaluated using the same annotations, the
	relative AP$_{75}$ difference cannot be explained solely by the diffuse
	nature of plume boundaries. Strict high-IoU localization therefore remains
	a limitation of the current model. In ECAF-Det, edge awareness refers to the
	use of boundary-sensitive structural information to support plume feature
	representation and detection; the present results do not indicate improved
	high-IoU localization.}

{To examine whether the large AP$_{50}$--AP$_{75}$ gap is associated with
	systematically oversized predictions, we further analyzed the geometry of
	one-to-one predicted and ground-truth boxes matched at IoU $\geq 0.5$.
	This conditional analysis directly examines the detections that satisfy the
	looser overlap criterion discussed above. The normalized center displacement
	is defined as the Euclidean distance between the predicted and ground-truth
	box centers divided by the diagonal length of the corresponding
	ground-truth box.}

\begin{table}[htbp]
	\centering
	\caption{Bounding-box geometry statistics for ECAF-Det on the IIG
		test set. Statistics are computed for one-to-one matched detections
		with IoU $\geq 0.5$.}
	\label{tab:bbox_geometry}
	\begin{tabular}{lc}
		\toprule
		Metric & Value \\
		\midrule
		Matched pairs & 1062 \\
		Pred./GT area ratio, median (IQR) & 0.912 (0.788--1.136) \\
		Pred./GT width ratio, median (IQR) & 0.960 (0.891--1.065) \\
		Pred./GT height ratio, median (IQR) & 0.960 (0.860--1.089) \\
		Normalized center displacement, median (IQR)
		& 0.080 (0.051--0.116) \\
		Area ratio $>1.25$ (\%) & 17.23 \\
		Area ratio $<0.80$ (\%) & 27.02 \\
		\bottomrule
	\end{tabular}
\end{table}

{As shown in Table~\ref{tab:bbox_geometry}, the median
	predicted-to-ground-truth area ratio is 0.912, while the median width and
	height ratios are both approximately 0.96. Moreover, 17.23\% of the
	matched predictions have an area more than 1.25 times the corresponding
	ground-truth box, whereas 27.02\% have an area below 0.80 times the
	ground-truth area. These results do not indicate a systematic tendency to
	produce overly expansive bounding boxes. Instead, the matched detections
	show a modest overall tendency toward under-coverage, together with residual
	center-alignment error.
	This analysis is conditional on detections already matched at
	IoU $\geq 0.5$ and therefore should not be interpreted as the distribution
	of all model predictions or as a substitute for AP$_{75}$. Rather, it
	directly tests whether detections satisfying the looser IoU criterion are
	predominantly obtained through excessive box expansion. The results suggest that the observed AP$_{50}$--AP$_{75}$ gap is more
	consistent with imperfect estimation of plume extent and residual spatial
	alignment than with systematic prediction-box enlargement.
	From an industrial application perspective, the AP$_{50}$--AP$_{75}$ gap
	also clarifies the current operating scope of ECAF-Det. The detector is
	suitable for identifying weak gas plumes and providing approximate
	regions of interest for early warning and subsequent inspection. However,
	the current high-IoU localization performance does not support precise
	leak-source localization from the bounding box alone. Applications that
	require accurate identification of the physical leakage origin would
	therefore require additional localization or source-estimation mechanisms.}

To further examine the performance of ECAF-Det on another infrared gas plume benchmark, we also evaluated it on the LangGas dataset. 
As shown in Table~\ref{tab:comparison_mainstream_LangGas}, ECAF-Det achieves an AP of 36.3\% and an AP$_{50}$ of 68.5\%, outperforming the compared detectors under the same experimental setting. 
Compared with RT-DETR-R18, ECAF-Det improves AP by 4.9 percentage points and AP$_{50}$ by 4.9 percentage points. 
{ECAF-Det also obtains the highest AP$_M$ and AP$_L$ among the evaluated methods, whereas RT-DETR-R34 achieves the highest AP$_{75}$ and AP$_S$ on this dataset.}
These results suggest that the proposed feature enhancement and adaptive fusion strategy is also effective on LangGas, although further cross-dataset evaluation is still needed to fully assess generalization across different imaging conditions and data sources.

\begin{figure}
	\centering
	\includegraphics[width=1\textwidth]{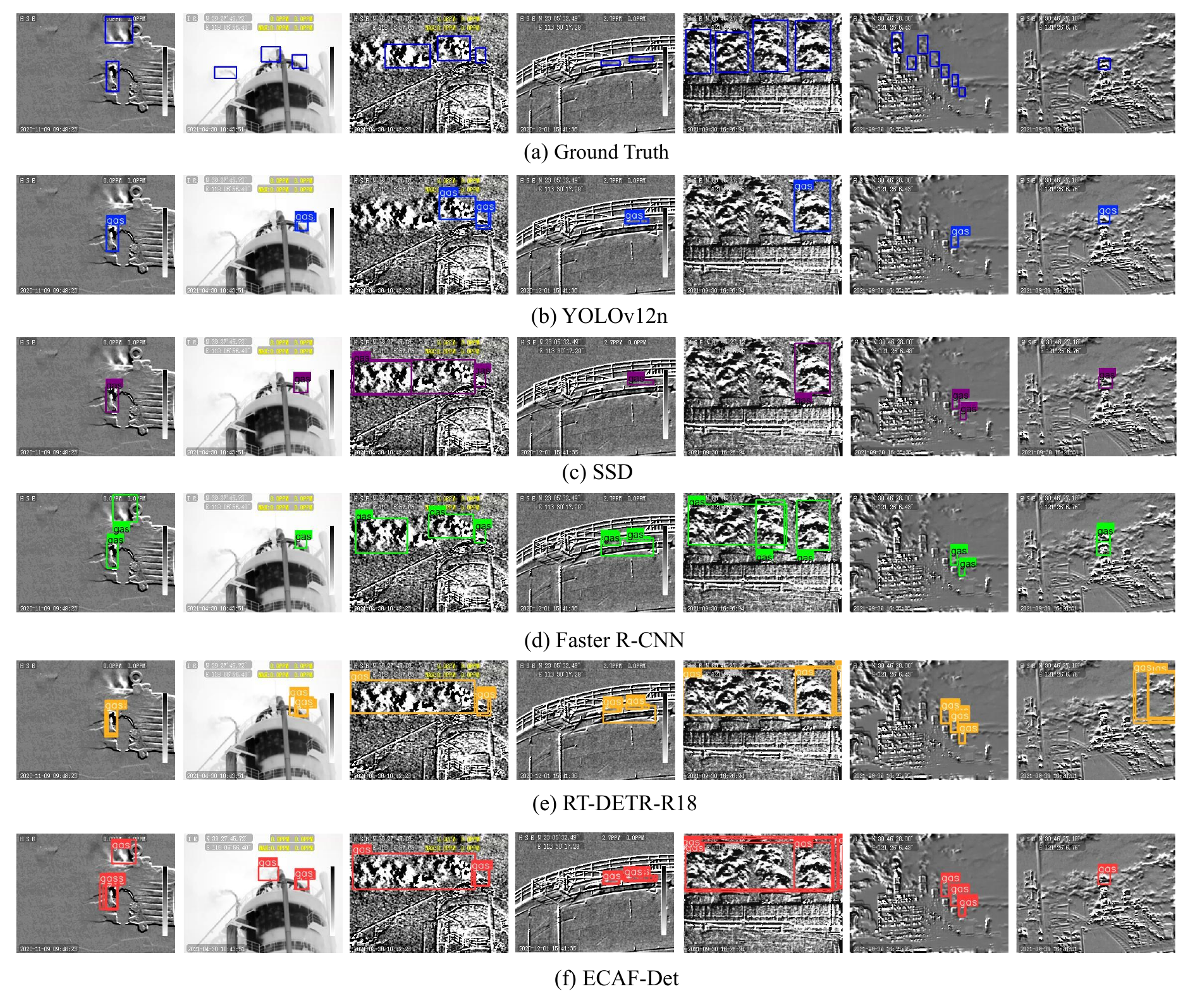}
	\caption{Qualitative comparison of detection results on the IIG dataset.
		All examples are original thermal infrared samples from the released IIG
		dataset. Each row shows the results of a different detector. ECAF-Det
		detects more weak and small plume regions in several challenging cases, but
		extremely faint and distant plumes remain difficult to localize accurately.}
	\label{fig:6}
\end{figure}

Qualitative detection results on the IIG dataset are shown in Figure~\ref{fig:6}. 
YOLOv12n is selected as the representative YOLO model because it achieves the best AP$_{50}$ among the evaluated YOLO variants on the IIG dataset. 
The visual comparison shows that lightweight one-stage detectors can miss faint or small plume regions under low-contrast conditions. 
SSD exhibits similar limitations in several examples, indicating that dense prediction alone is insufficient when plume boundaries are weak and background thermal patterns are complex. 
Faster R-CNN can detect some plume regions more accurately, but its computational cost is substantially higher. 
RT-DETR-R18 provides a better balance between accuracy and efficiency, but still produces missed detections or incomplete boxes in challenging scenes.

Compared with these detectors, ECAF-Det produces more complete detections in several weak-plume cases and better preserves small plume regions. 
However, failure cases still occur when gas plumes are extremely small, visually indistinct, or located far from the camera. 
Detection is also more difficult in cluttered thermal backgrounds than in simple and uniform scenes. 
These observations indicate that weak-plume localization under complex industrial backgrounds remains challenging and should be further investigated in future work.

\noindent{\textbf{Challenge-oriented evaluation.}}
{
	To further quantify detection performance under challenging plume and
	background conditions, we performed a post-hoc analysis on the IIG
	test set. No additional model training or parameter optimization was
	conducted. Instead, the existing predictions of RT-DETR-R18 and ECAF-Det were
	re-evaluated after the IIG test set was stratified according to
	image-based plume visibility and local thermal-background complexity.
}
{
	For each annotated plume instance, the infrared image was first robustly
	normalized to $[0,1]$ using its 1st and 99th intensity percentiles.
	A local background region was then constructed by expanding the
	ground-truth bounding box to 1.5 times its original width and height and
	excluding the original box. The visibility of the $j$-th plume instance
	in image $i$ was characterized using the local contrast-to-noise ratio		
	\begin{equation}
		V_{i,j} =
		\frac{|\mu_{p,i,j}-\mu_{b,i,j}|}
		{\sigma_{b,i,j}+\epsilon},
	\end{equation}	
	where $\mu_{p,i,j}$ denotes the mean normalized infrared intensity within
	the annotated plume box, while $\mu_{b,i,j}$ and $\sigma_{b,i,j}$ denote
	the mean and standard deviation of the surrounding local background,
	respectively. For images containing multiple annotated plume instances,
	the instance-level visibility scores were aggregated using their median,
	i.e., $V_i=\operatorname{median}_{j}V_{i,j}$, to obtain a single
	image-level visibility score. The test images were then ranked according
	to $V_i$ and divided into approximately equal weak-, moderate-, and
	strong-visibility groups using the 33.3rd and 66.7th percentiles.
	The corresponding thresholds were 0.099 and 0.239.}
{
	The resulting score characterizes the visual separability of a plume from
	its local thermal background and is not interpreted as a physical
	measurement of gas concentration or leakage rate.
}
{
	Local thermal-background complexity was quantified for each annotated
	plume instance using the mean Sobel gradient magnitude within its
	surrounding local background region. For images containing multiple
	annotated plume instances, the instance-level complexity scores were
	aggregated using their median to obtain a single image-level background
	complexity score. The test images were then divided into approximately
	equal simple-, moderate-, and complex-background groups according to the
	33.3rd and 66.7th percentiles of the image-level scores.
}
\begin{table}[htbp]
	\centering
	\caption{{Challenge-oriented detection performance on the IIG test set.}}
	\label{tab:challenge_analysis}
	\scriptsize
	\resizebox{0.6\columnwidth}{!}{
		\begin{tabular}{llcccc}
			\toprule
			Condition & Model & AP (\%) & AP$_{50}$ (\%) & AP$_{75}$ (\%) & AR$_{100}$ (\%) \\
			\midrule
			
			\multicolumn{6}{c}{\textit{Plume visibility}} \\
			\midrule
			
			Weak
			& RT-DETR-R18 & 24.8 & 75.4 & 4.5 & 51.2 \\
			& ECAF-Det
			& \textbf{29.5}
			& \textbf{88.0}
			& \textbf{4.9}
			& \textbf{53.8} \\
			
			Moderate
			& RT-DETR-R18 & 26.9 & 79.1 & 9.3 & 53.9 \\
			& ECAF-Det
			& \textbf{32.1}
			& \textbf{87.2}
			& \textbf{11.9}
			& \textbf{56.5} \\
			
			Strong
			& RT-DETR-R18 & 28.4 & 74.5 & \textbf{12.1} & 49.8 \\
			& ECAF-Det
			& \textbf{29.2}
			& \textbf{77.9}
			& 10.9
			& \textbf{53.3} \\
			
			\midrule
			
			\multicolumn{6}{c}{\textit{Local thermal-background complexity}} \\
			\midrule
			
			Simple
			& RT-DETR-R18
			& \textbf{30.5}
			& \textbf{83.2}
			& \textbf{9.7}
			& \textbf{54.0} \\
			& ECAF-Det
			& 27.8
			& 80.7
			& 9.1
			& 50.8 \\
			
			Moderate
			& RT-DETR-R18 & 34.1 & 86.7 & 18.2 & 58.0 \\
			& ECAF-Det
			& \textbf{36.6}
			& \textbf{88.0}
			& \textbf{19.1}
			& \textbf{59.2} \\
			
			Complex
			& RT-DETR-R18 & 18.0 & 60.5 & 1.8 & 42.9 \\
			& ECAF-Det
			& \textbf{27.5}
			& \textbf{85.6}
			& \textbf{3.1}
			& \textbf{53.5} \\
			
			\bottomrule
		\end{tabular}
	}
\end{table}
{
	As shown in Table~\ref{tab:challenge_analysis}, ECAF-Det provides larger
	gains when plume visibility is limited. In the weak-visibility group,
	ECAF-Det improves AP from 24.8\% to 29.5\% and AP$_{50}$ from 75.4\%
	to 88.0\% relative to RT-DETR-R18, corresponding to gains of 4.7 and
	12.6 percentage points, respectively. For moderate-visibility plumes, AP
	increases from 26.9\% to 32.1\%, whereas the improvement is reduced to
	0.8 percentage points for strong-visibility plumes. These results indicate
	that the performance benefit is concentrated more strongly in low- and
	moderate-visibility plume scenes than in visually salient scenes.
}
{
	The difference becomes more pronounced as local thermal-background
	complexity increases. Under complex thermal backgrounds, RT-DETR-R18
	obtains an AP of 18.0\% and an AP$_{50}$ of 60.5\%, whereas ECAF-Det
	reaches 27.5\% and 85.6\%, corresponding to improvements of 9.5 and
	25.1 percentage points, respectively. AR$_{100}$ also increases from
	42.9\% to 53.5\%. These results show that ECAF-Det is substantially more
	robust when weak plume responses coexist with strong local thermal
	structures.
}
{
	The improvement is not uniform across all scene conditions. On the
	simple-background subset, ECAF-Det obtains an AP of 27.8\%, compared
	with 30.5\% for RT-DETR-R18. This suggests that the proposed task-oriented
	feature enhancement and adaptive fusion provide limited additional benefit
	when the local thermal background is already relatively simple and may
	introduce a modest trade-off in such easier scenes. Therefore, the
	challenge-oriented results should not be interpreted as uniform superiority
	under all conditions; rather, the main advantage of ECAF-Det is concentrated
	in weak-plume and thermally cluttered scenes.
}
{
	Together with the existing AP$_S$ result, where ECAF-Det improves
	small-plume AP from 19.9\% to 25.3\%, the challenge-oriented analysis
	provides complementary quantitative evidence for weak-visibility,
	small-plume, and complex thermal-background conditions.
}

To complement the standard COCO metrics, we report additional
detection-oriented metrics on the IIG test set in Table~\ref{tab:industrial_alarm_metrics}. These are image-level detection metrics and should not be interpreted
as event-level false-alarm statistics. Unlike standard COCO metrics, these indicators provide complementary
information on detection reliability and missed-detection behavior. Precision reflects the reliability of generated alarms, recall
indicates the ability to detect leakage instances, and the missed-detection
rate is calculated as \(1-\mathrm{Recall}\). {In addition, model-inference latency and end-to-end FPS are reported to
	characterize the computational responsiveness of the detector on the evaluated
	RTX 5070 Ti workstation GPU. These measurements provide a platform-specific
	assessment of inference efficiency rather than a direct evaluation of real-time
	performance on low-power industrial edge hardware.}

As shown in Table~\ref{tab:industrial_alarm_metrics}, ECAF-Det improves the
recall from 70.60\% to 79.44\% compared with RT-DETR-R18, corresponding to
an increase of 8.84 percentage points. Meanwhile, the missed-detection rate
is reduced from 29.40\% to 20.56\%. This result is important for engineering deployment because missed detections
of weak early-stage plumes may reduce the reliability of automated visual
inspection systems and delay operator response. The F1-score is
also improved from 77.94\% to 82.73\%, indicating that ECAF-Det achieves a
better balance between alarm reliability and leakage detectability.

Although the precision of ECAF-Det is slightly lower than that of
RT-DETR-R18, decreasing from 86.97\% to 86.31\%, the decrease is marginal
compared with the substantial improvement in recall. From an engineering application perspective, this trade-off is acceptable
because reducing missed weak plumes is often more critical than obtaining a
small gain in precision for automated gas-leak monitoring. {
	In terms of computational efficiency, ECAF-Det requires an average model-inference time of 15.46 ms per image and achieves an end-to-end throughput of 61.81 FPS on the NVIDIA GeForce RTX 5070 Ti used in our experimental platform, with a batch size of 4.
	These results demonstrate an end-to-end throughput of 61.81 FPS on the evaluated workstation GPU while providing stronger weak-plume detection performance. However, since inference speed is strongly hardware-dependent, the current measurements should not be generalized to low-power edge devices, and dedicated edge-platform evaluation remains necessary before practical deployment on such hardware.}

\subsection{ Ablation Experiments }

\begin{table}[htbp]
	\centering
	\caption{Ablation study of the plume-oriented local--global backbone
		on the IIG dataset. The local--global plume block denotes the
		proposed block for local detail preservation and frequency-domain
		contextual aggregation. MSEPM denotes the multi-scale edge perception
		module using GGEO to generate hierarchical boundary-sensitive
		structural priors.}
	\label{tab:ablation_backbone}
	\resizebox{\textwidth}{!}{
		\begin{tabular}{ccccccccc}
			\toprule
			Local--Global Plume Block & MSEPM &
			AP & AP$_{50}$  &
			AP$_S$ & AP$_M$ & AP$_L$ &
			GFLOPs (G) & Params (M) \\
			\midrule
			$\times$ & $\times$
			& 26.8 & 77.8 
			& 19.9 & 32.0 & 37.1 & 56.9 & 19.873 \\
			
			$\checkmark$ & $\times$
			& 28.5 & 80.7 
			& 22.2 & 32.8 & 37.3 & 51.9 & 15.965 \\
			
			$\times$ & $\checkmark$
			& 27.9 & 80.4 
			& 24.0 & 30.6 & 38.1 & 59.8 & 21.119 \\
			
			$\checkmark$ & $\checkmark$
			& \textbf{29.5} & \textbf{81.0} 
			& \textbf{24.3} & \textbf{33.5} & \textbf{38.9}
			& 54.8 & 17.211 \\
			\bottomrule
		\end{tabular}
	}
\end{table}

{To evaluate the contribution of the proposed backbone components, we conducted ablation experiments by
	progressively introducing the local--global plume block and MSEPM into the RT-DETR-R18 baseline. The results are reported in Table~\ref{tab:ablation_backbone}. The baseline achieves an AP of 26.8\% and an AP$_{50}$ of 77.8\%. Introducing the local--global plume block alone increases AP to 28.5\%
	and AP$_{50}$ to 80.7\%. At the same time,
	the computational cost decreases from 56.9 to 51.9 GFLOPs. These results indicate that the
	local--global design primarily improves overall plume detection and contextual representation.}
{
	A controlled replacement analysis of the global-context branch is further
	provided in Section~4.5.11 to determine whether the observed improvement
	depends specifically on the proposed DCT-based frequency-domain modeling
	or can be reproduced by alternative spatial-domain global-context
	mechanisms.
}
{
	Using MSEPM alone improves AP from 26.8\% to 27.9\% and AP$_S$
	from 19.9\% to 24.0\%. Despite the absence
	of direct pixel-level edge supervision, these improvements indicate that the
	boundary-sensitive structural priors provide task-useful information for plume detection,
	particularly for small plumes. This result
	should be interpreted as evidence of detection-level utility rather than proof that the
	generated edge maps constitute exact semantic plume boundaries. However, MSEPM also
	increases the computational cost to 59.8 GFLOPs, indicating that its accuracy gain should
	be considered together with its additional overhead. A controlled edge-prior randomization
	and replacement analysis is further provided in Section~4.5.8 to distinguish the contribution
	of meaningful structural priors from that of generic feature perturbation.}

{When the local--global plume block and MSEPM are combined, the backbone achieves the best overall performance
	in Table~\ref{tab:ablation_backbone}. More specifically, relative to the local--global plume block alone,
	introducing MSEPM increases AP from 28.5\% to 29.5\%, AP$_{50}$ from
	80.7\% to 81.0\%, AP$_S$ from 22.2\% to 24.3\%, AP$_M$ from
	32.8\% to 33.5\%, and AP$_L$ from 37.3\% to 38.9\%. Thus, the additional contributions of MSEPM in the intended backbone
	configuration are +1.0, +0.3, +2.1, +0.7, and +1.6 percentage points
	for AP, AP$_{50}$, AP$_S$, AP$_M$, and AP$_L$, respectively. These results further support interpreting MSEPM
	as a complementary boundary-sensitive structural-prior pathway rather than as an
	independent performance-driving component. The local--global plume block provides the
	main local-detail and contextual representation, while MSEPM supplies additional structural
	guidance. Its largest complementary gain is observed for small plumes, which is consistent
	with the objective of preserving weak and easily attenuated plume structures. This
	improvement is obtained with an increase in computational cost from 51.9 to 54.8 GFLOPs
	and in parameter count from 15.965 M to 17.211 M. Therefore, the additional 2.9 GFLOPs
	should be interpreted against the combined-backbone gains, particularly the
	2.1-percentage-point improvement in AP$_S$, rather than against overall AP alone.}

\begin{table}[htbp]
	\centering
	\caption{Ablation study of the complete ECAF-Det on the IIG dataset.}
	\label{tab:ablation_complete}
	\resizebox{\textwidth}{!}{
		\begin{tabular}{ccccccccc}
			\toprule
			Plume-Oriented Local--Global Backbone & CASR-PAN &
			AP & AP$_{50}$  &
			AP$_S$ & AP$_M$ & AP$_L$ &
			GFLOPs (G) & Params (M) \\
			\midrule
			$\times$ & $\times$
			& 26.8 & 77.8 
			& 19.9 & 32.0 & 37.1 & 56.9 & 19.873 \\
			
			$\checkmark$ & $\times$
			& 29.5 & 81.0 
			& 24.3 & 33.5 & 38.9 & 54.8 & 17.211 \\
			
			$\times$ & $\checkmark$
			& 29.4 & 79.1 
			& 23.0 & \textbf{33.6} & 42.2 & 45.8 & 16.939 \\
			
			$\checkmark$ & $\checkmark$
			& \textbf{29.8} & \textbf{84.3} 
			& \textbf{25.3} & 32.5 & \textbf{42.6}
			& \textbf{43.7} & \textbf{14.277} \\
			\bottomrule
		\end{tabular}
	}
\end{table}

{\color{red}
	Table~\ref{tab:ablation_complete} reports the ablation results of the
	complete ECAF-Det. Replacing the original RT-DETR-R18 backbone with the
	proposed plume-oriented local--global backbone increases AP from 26.8\%
	to 29.5\%, AP$_{50}$ from 77.8\% to 81.0\%, and AP$_S$ from 19.9\%
	to 24.3\%, while reducing the computational cost from 56.9 to
	54.8 GFLOPs and the parameter count from 19.873 M to 17.211 M.
	Using CASR-PAN alone also provides a substantial improvement, increasing
	AP to 29.4\% and AP$_L$ to 42.2\%, while reducing the computational
	cost to 45.8 GFLOPs and the parameter count to 16.939 M.
	Combining the plume-oriented local--global backbone with CASR-PAN yields
	the best overall AP of 29.8\%, together with the highest AP$_{50}$ of
	84.3\%, AP$_S$ of 25.3\%, and AP$_L$ of 42.6\%. The complete model
	also requires only 43.7 GFLOPs and 14.277 M parameters. These results
	demonstrate that the two components provide complementary benefits for
	weak-plume representation and adaptive multi-scale feature aggregation,
	resulting in the strongest overall accuracy--efficiency balance among
	the evaluated configurations.
}

\subsection{More Experiments}

\subsubsection{Stage-wise Placement of Local--Global Plume Blocks}

{
	To determine the appropriate placement of the local--global plume block,
	we replaced ResNet-18 BasicBlocks at different backbone stages and compared
	the resulting accuracy--complexity trade-offs. This experiment evaluates
	whether local--global enhancement is more effective when introduced at
	shallow, deep, or multiple feature stages. The evaluated configurations are summarized in
	Table~\ref{tab:placement_lgpb}.	}		
\begin{table}[t]
	\centering
	\caption{Stage-wise placement of local--global plume blocks in the ResNet-18 backbone on the IIG dataset.}
	\label{tab:placement_lgpb}
	\scriptsize
	\setlength{\tabcolsep}{2.5pt}
	\renewcommand{\arraystretch}{1.1}
	\resizebox{\columnwidth}{!}{
		\begin{tabular}{lccccccc}
			\toprule
			Configuration & AP & AP$_{50}$ & AP$_S$ & AP$_M$ & AP$_L$ & GFLOPs & Params \\
			\midrule
			All BasicBlocks baseline     
			& 26.8 & 77.8 & 19.9 & 32.0 & 37.1 & 56.9 & 19.873 \\
			
			P2--P5 local--global plume blocks           
			& 27.4 & 77.9 & 21.1 & 31.1 & 35.7 & \textbf{47.1} & \textbf{15.725} \\
			
			P4--P5 local--global plume blocks 
			& \textbf{28.5} & \textbf{80.7} & 22.2 & \textbf{32.8} & 37.3 & 51.9 & 15.965 \\
			
			P5 local--global plume block     
			& 27.0 & 78.6 & \textbf{22.4} & 30.2 & \textbf{42.8} & 54.5 & 16.743 \\
			\bottomrule
		\end{tabular}
	}
\end{table}		
{
	As shown in Table~\ref{tab:placement_lgpb}, the P4--P5 configuration achieves the best overall
	performance, with an AP of 28.5\% and an AP$_{50}$ of 80.7\%, while
	reducing the computational cost from 56.9 to 51.9 GFLOPs and the parameter
	count from 19.873 M to 15.965 M. Although replacing all stages from P2 to
	P5 further reduces computational cost, its AP is lower at 27.4\%. Using
	the local--global plume block only at P5 improves AP$_S$ and AP$_L$, but
	provides limited improvement in overall AP.}		
{
	Based on these results, the P4--P5 configuration is adopted in ECAF-Det
	because it provides the best overall accuracy among the evaluated placement
	strategies while maintaining moderate computational cost.	}

\subsubsection{Effective Receptive Field Analysis of the Plume-Oriented Local--Global Backbone}
\begin{figure}
	\centering
	\includegraphics[width=0.8\textwidth]{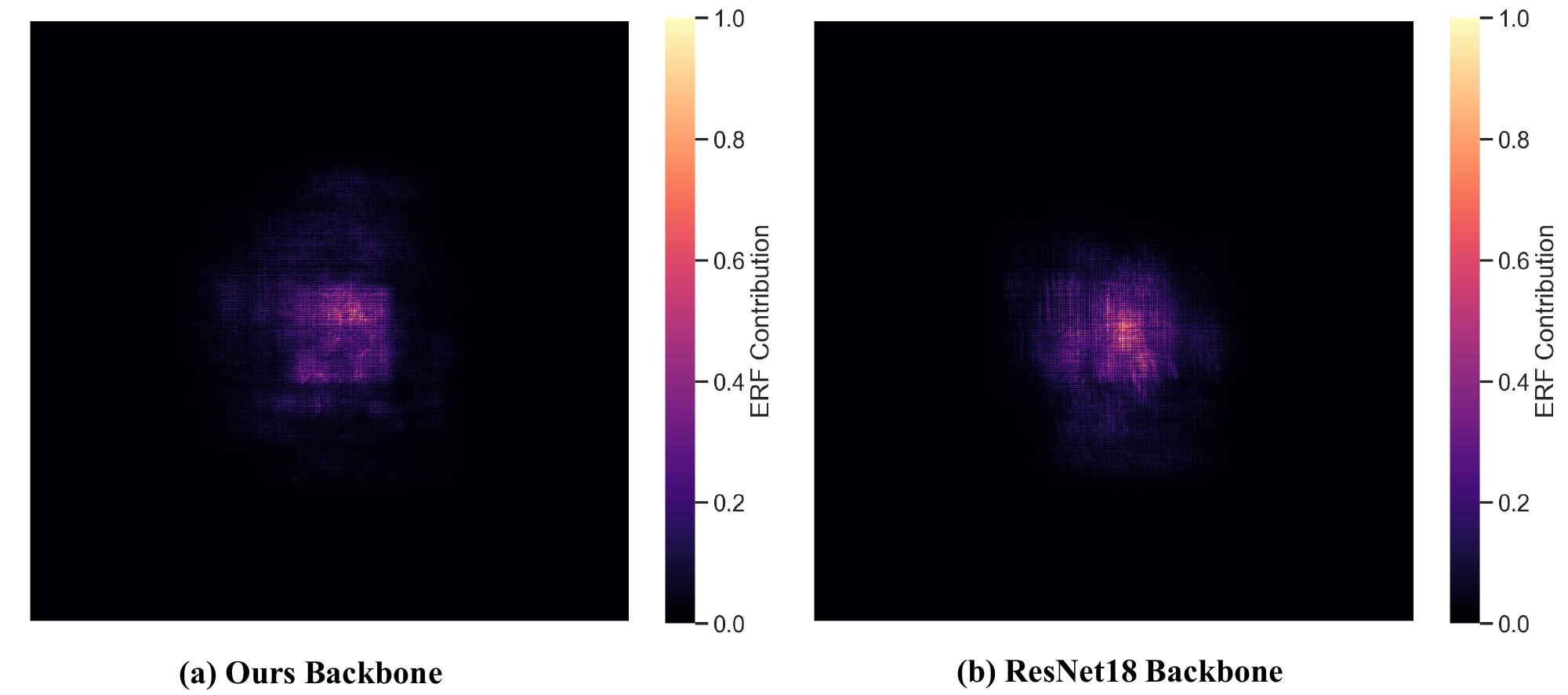}
	\caption{Effective receptive field (ERF) visualization comparison between 
		(a) the proposed plume-oriented local--global backbone and 
		(b) the original ResNet-18 backbone under the same training and inference settings.}
	\label{fig:7}
\end{figure}

\begin{figure}  
	\centering
	\includegraphics[width=0.6\textwidth]{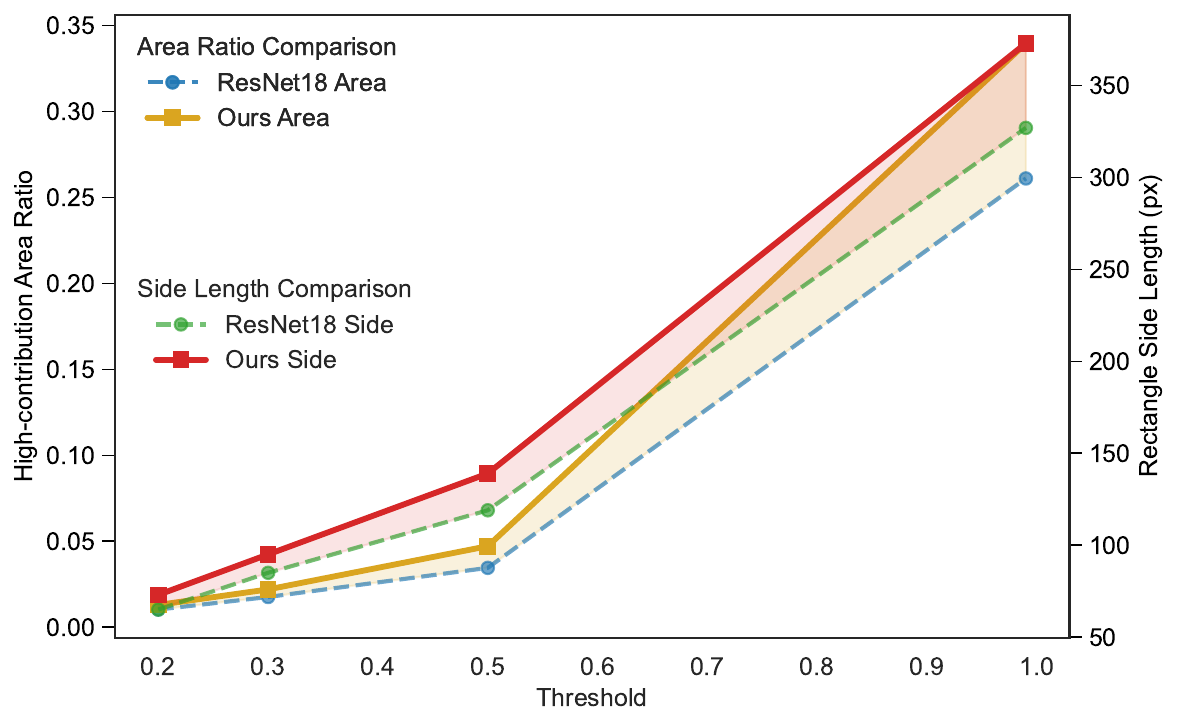} 
	\caption{Comparison of high-contribution area ratios between the proposed plume-oriented local--global backbone and the original ResNet-18 backbone under different cumulative contribution thresholds.}
	\label{fig:8}
\end{figure}

{
	To evaluate the contextual aggregation ability of the proposed backbone,
	we performed effective receptive field (ERF) analysis following~\citep{ding2022scaling}. The ERF heatmap is obtained by backpropagating
	the central activation to the input and accumulating the corresponding
	gradient responses.
	As shown in Figure~\ref{fig:7}, the proposed backbone produces a broader and more
	spatially continuous ERF than the original ResNet-18 backbone.
	More specifically, the high-response region of the proposed backbone extends
	farther from the central activation and exhibits a more continuous spatial
	distribution, whereas the response of the ResNet-18 backbone is more
	concentrated around the center. This qualitative difference suggests that the
	proposed local--global backbone maintains sensitivity to a wider spatial
	context, which is beneficial for representing diffuse gas plumes whose
	discriminative cues may extend beyond a compact local neighborhood.
	Figure~\ref{fig:8} provides quantitative support through the
	high-contribution area ratio.
	At cumulative contribution thresholds of 0.5 and 0.99, the proposed
	backbone obtains area ratios of 0.047 and 0.340, compared with 0.036 and
	0.255 for ResNet-18, respectively. These results support the stronger
	long-range contextual aggregation of the proposed backbone.
	ERF analysis reflects feature sensitivity rather than physical gas transport
	behavior and is therefore used only as supporting evidence for contextual
	aggregation ability.
}

\begin{figure}
	\centering
	\includegraphics[width=\textwidth]{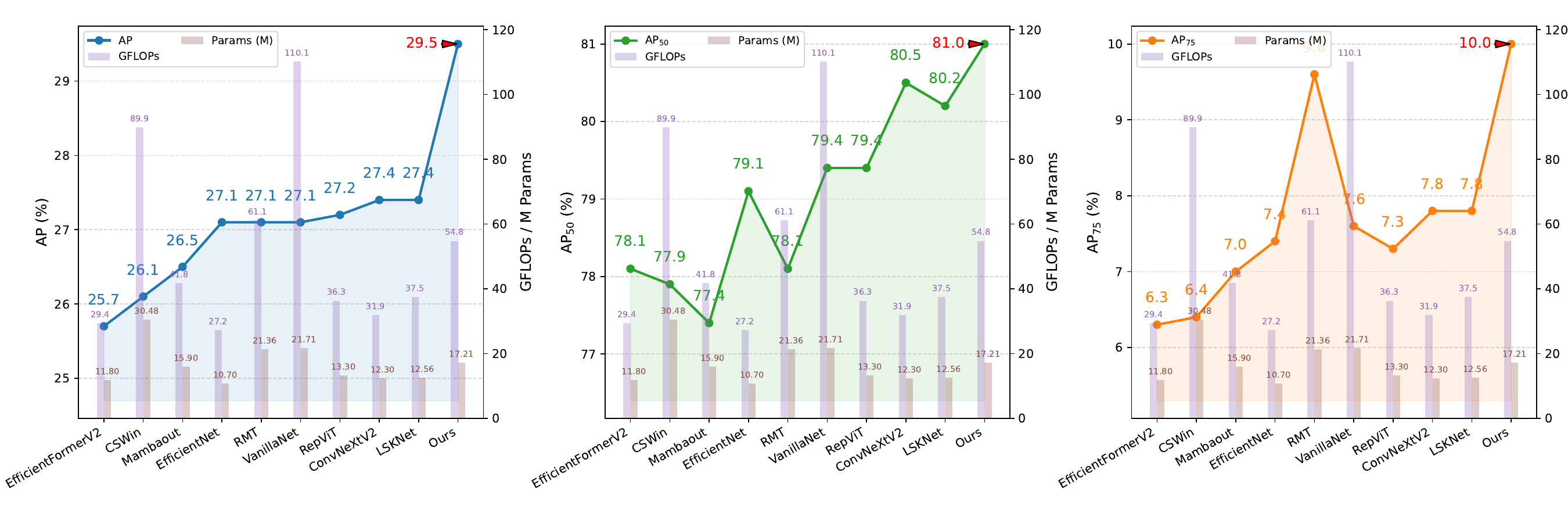}
	\caption{Performance comparison of different backbone architectures on the IIG dataset. 
		The marked point denotes the proposed plume-oriented local--global backbone.}
	\label{fig:9}
\end{figure}

\subsubsection{Backbone-Level Comparison with Representative Architectures}

To further evaluate the feature extraction capability of the proposed backbone, we compared it with several representative backbone architectures, including EfficientNet~\citep{tan2019efficientnet}, MambaOut~\citep{yu2025mambaout}, RepViT~\citep{wang2024repvit}, CSWin Transformer~\citep{dong2022cswin}, ConvNeXtV2~\citep{woo2023convnext}, RMT~\citep{fan2024rmt}, EfficientFormerV2~\citep{li2023rethinking}, LSKNet~\citep{li2023large}, and VanillaNet~\citep{chen2023vanillanet}. 
All backbones were evaluated under the same detection framework and training settings, and the results are summarized in Figure~\ref{fig:9}.

The proposed plume-oriented local--global backbone achieves the best overall performance among the evaluated backbones, with an AP of 29.5\%, an AP$_{50}$ of 81.0\%, and an AP$_{75}$ of 10.0\%. 
Compared with lightweight convolutional backbones such as EfficientNet and RepViT, the proposed backbone obtains higher AP with a comparable parameter scale. 
This suggests that combining local detail preservation with long-range contextual aggregation is beneficial for detecting faint and diffuse gas plumes.

Compared with transformer-based or large-kernel architectures such as CSWin Transformer, RMT, and LSKNet, the proposed backbone achieves competitive or better detection accuracy while maintaining moderate computational complexity. 
This indicates that the proposed design provides a favorable balance between detection performance and model cost for infrared gas leak detection. 
The result is also consistent with the ERF analysis in Figures~\ref{fig:7} and~\ref{fig:8}, where the proposed backbone shows a broader effective receptive field and stronger spatial continuity than the original ResNet-18 backbone.

Overall, the backbone-level comparison demonstrates that the proposed plume-oriented local--global backbone is effective for weak infrared gas plume representation. 
However, this comparison is conducted within the current detection framework and dataset setting; further validation on additional industrial datasets and deployment platforms is still needed to assess broader generalization.

\begin{figure}
	\centering
	\includegraphics[width=\textwidth]{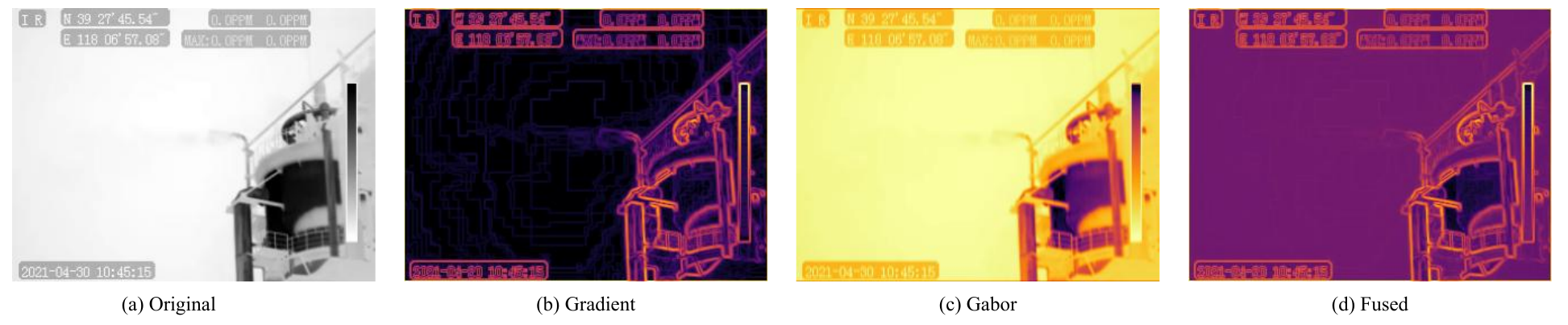}
	\caption{
		Visualization results of the proposed gradient--Gabor edge operator (GGEO).
		(a) Original infrared image;
		(b) directional-gradient response extracted by the fixed Sobel-like operators;
		(c) Gabor-response cue obtained from the depthwise Gabor branch;
		(d) fused structural response combining directional-gradient and Gabor cues.
	}
	\label{fig:10}
\end{figure}

\subsubsection{Visualization Analysis of GGEO Structural Representations}

To further analyze the structural representation produced by the proposed
gradient--Gabor edge operator (GGEO), Figure~\ref{fig:10} visualizes the
original infrared image, the directional-gradient response, the Gabor-response
cue, and the fused structural response. 
The gradient branch, implemented using multi-directional Sobel operators, highlights local intensity transitions and captures prominent structural contours in the scene. 
However, because infrared images often contain background temperature variations and thermal noise, the gradient cue also responds to some non-plume structures, leading to fragmented or redundant edge responses in background regions.

The Gabor-response branch, implemented using Gabor-based filtering, provides a smoother structural response in weak-edge regions. 
Compared with the gradient cue, the Gabor-response cue shows better spatial continuity and is less sensitive to isolated intensity fluctuations. 
This property is useful for semi-transparent gas plumes, whose boundaries are often diffuse and only weakly contrasted against the background.

The fused structural response combines the complementary strengths of the two branches. 
It retains the local boundary sensitivity of the gradient cue while improving the continuity of weak plume-related structures through the gabor-response cue. 
As shown in Figure~\ref{fig:10}(d), the fused edge representation makes the plume region more distinguishable from the surrounding thermal background in several weak-boundary areas. 
These visualization results provide qualitative evidence that the proposed edge representation can supply useful boundary-aware auxiliary information for subsequent plume feature extraction and multi-scale fusion.

{
	Because GGEO is not trained with pixel-level edge annotations, the responses
	in Figure~\ref{fig:10} represent boundary-sensitive structural cues rather
	than semantic plume-edge predictions. Residual responses to background
	thermal structures are therefore expected. Their task-level usefulness is
	evaluated through the detection results and ablation studies rather than
	through direct edge-map accuracy.
}

\begin{figure}
	\centering
	\includegraphics[width=\textwidth]{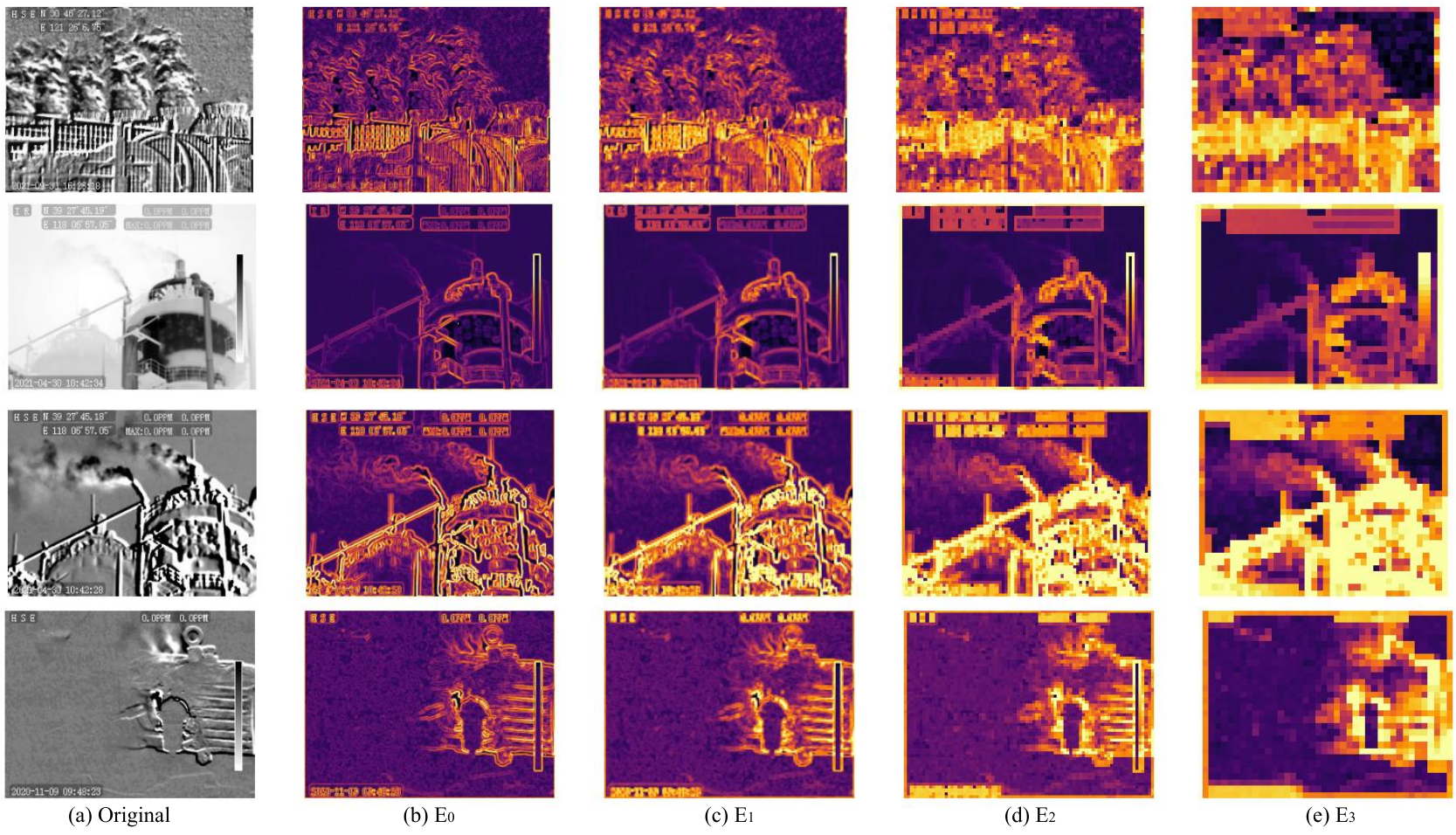}
	\caption{
		Visualization of the hierarchical structural representations involved in
		the multi-scale edge perception module (MSEPM).
		(a) Original infrared image;
		(b) $E_0$: the initial structural representation generated by GGEO before
		downsampling;
		(c) $E_1$: the representation after one $2\times2$ max-pooling operation;
		(d) $E_2$: the representation after two successive max-pooling operations;
		and
		(e) $E_3$: the representation after three successive max-pooling operations.
		The downsampled representations $E_1$, $E_2$, and $E_3$ are subsequently
		projected through scale-specific $1\times1$ convolutions to form the
		hierarchical structural priors used by the corresponding backbone stages.
	}
	\label{fig:11}
\end{figure}

\subsubsection{Multi-Scale Structural Representation Analysis of MSEPM}

{
	To further analyze how structural information evolves through the proposed
	multi-scale edge perception module (MSEPM), Figure~\ref{fig:11} visualizes
	the initial GGEO output $E_0$ together with the three progressively
	downsampled representations $E_1$, $E_2$, and $E_3$. $E_0$ is the intermediate structural representation generated by GGEO and is
	not directly used as a backbone-stage output. According to
	Eqs.~(11)--(12), $E_1$, $E_2$, and $E_3$ are obtained through one, two,
	and three successive $2\times2$ max-pooling operations, respectively,
	and are subsequently projected by scale-specific $1\times1$ convolutions
	to form the hierarchical structural priors injected into the corresponding
	backbone stages.
}

{
	As shown in Figure~\ref{fig:11}(b), $E_0$ preserves the richest
	fine-grained structural information because it is generated directly by
	GGEO before spatial downsampling. Local intensity transitions and weak
	plume-related structures remain visible at this stage, together with some
	responses originating from background thermal textures. This observation
	is consistent with the fact that GGEO provides a boundary-sensitive
	structural representation rather than an explicitly supervised semantic
	plume-edge map.
}

{
	After one max-pooling operation, $E_1$ exhibits a more compact structural
	response while retaining relatively fine spatial information. Compared
	with $E_0$, some small-scale fluctuations are reduced, whereas the dominant
	plume-related structures remain visible. This representation therefore
	provides relatively detailed auxiliary information for the shallower
	backbone stage.
}

{
	The representation $E_2$, obtained after two successive pooling operations,
	further aggregates neighboring responses and emphasizes broader structural
	patterns. Fine local variations become less prominent, while the spatial
	continuity of the main plume region is retained. After the third pooling
	operation, $E_3$ mainly describes coarse structural information and the
	overall spatial extent of the plume region, providing a larger-scale
	structural cue for the deeper backbone stage.	}
{
	Overall, the visualization illustrates a progressive transition from
	fine-grained structural responses in $E_0$ to increasingly coarse and
	spatially aggregated representations in $E_1$, $E_2$, and $E_3$.
	In the implemented architecture, $E_0$ serves as the internal GGEO
	representation, whereas $E_1$, $E_2$, and $E_3$ are projected through
	scale-specific $1\times1$ convolutions to produce the hierarchical
	boundary-sensitive structural priors used by the backbone.	}

\subsubsection{Ablation on Gradient Directions in GGEO}

An ablation study was conducted to analyze the effect of gradient direction modeling in the proposed gradient--Gabor edge operator (GGEO), which is embedded in the multi-scale edge perception module (MSEPM). 
GGEO uses fixed Sobel-like directional kernels to extract orientation-sensitive edge responses, which are then combined with gabor-response cues. 
To evaluate the contribution of directional diversity, we compared three configurations: a single horizontal direction ($0^\circ$), two orthogonal directions ($0^\circ$ and $90^\circ$), and four directions ($0^\circ$, $45^\circ$, $90^\circ$, and $135^\circ$). 
All variants were implemented within the same MSEPM architecture and used identical downstream detection components. 
The only difference was the number of gradient orientations used in the gradient branch.

\begin{table}
	\centering
	\caption{Ablation study of gradient directions in GGEO on the IIG dataset.}
	\label{tab:ablation_GGEO}
	\resizebox{\textwidth}{!}{
		\begin{tabular}{lccccccccc}
			\toprule
			Gradient Directions & AP & AP$_{50}$ & AP$_{75}$ & AP$_S$ & AP$_M$ & AP$_L$ & GFLOPs (G) & Params (M) \\
			\midrule
			$0^\circ$ & 26.7 & 79.8 & 6.3 & 21.4 & 31.1 & 33.8 & \textbf{59.7} & \textbf{21.117} \\
			$0^\circ$, $90^\circ$ & \textbf{28.8} & \textbf{81.1} & 9.4 & 22.1 & \textbf{33.1} & \textbf{42.3} & \textbf{59.7} & 21.118 \\
			$0^\circ$, $45^\circ$, $90^\circ$, $135^\circ$ & 27.9 & 80.4 & \textbf{10.3} & \textbf{24.0} & 30.6 & 38.1 & 59.8 & 21.119 \\
			\bottomrule
		\end{tabular}
	}
\end{table}
\begin{figure}
	\centering
	\includegraphics[width=1\textwidth]{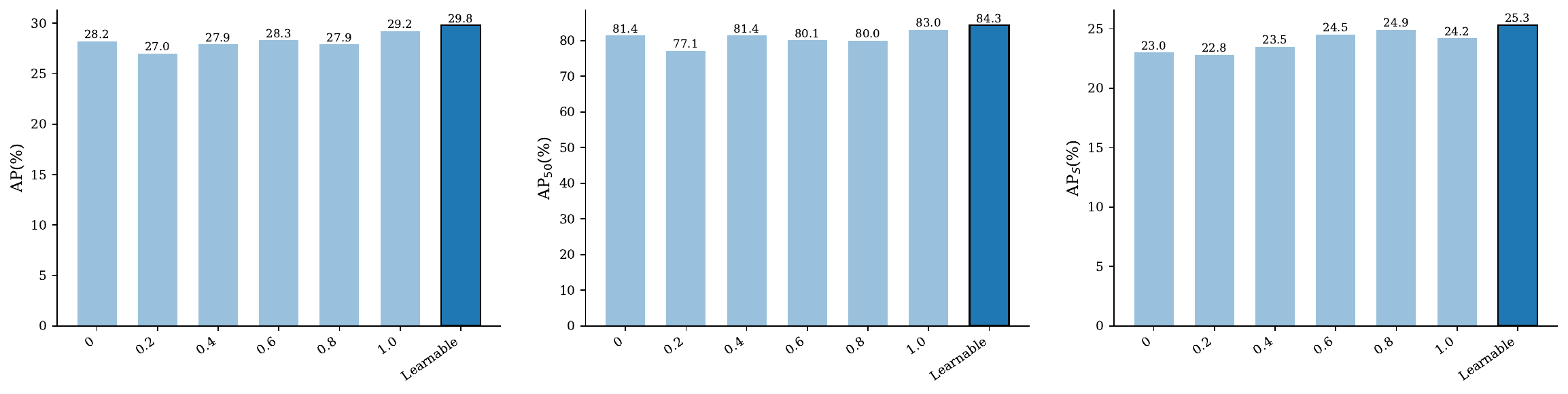}
	\caption{Effect of the fusion coefficient $\alpha$ in GGEO. 
		Performance comparisons under different fixed values of $\alpha$ and learnable $\alpha$ are reported in terms of 
		(a) AP, (b) AP$_{50}$, and (c) AP$_S$.}
	\label{fig:12}
\end{figure}
\begin{figure}
	\centering
	\includegraphics[width=0.5\textwidth]{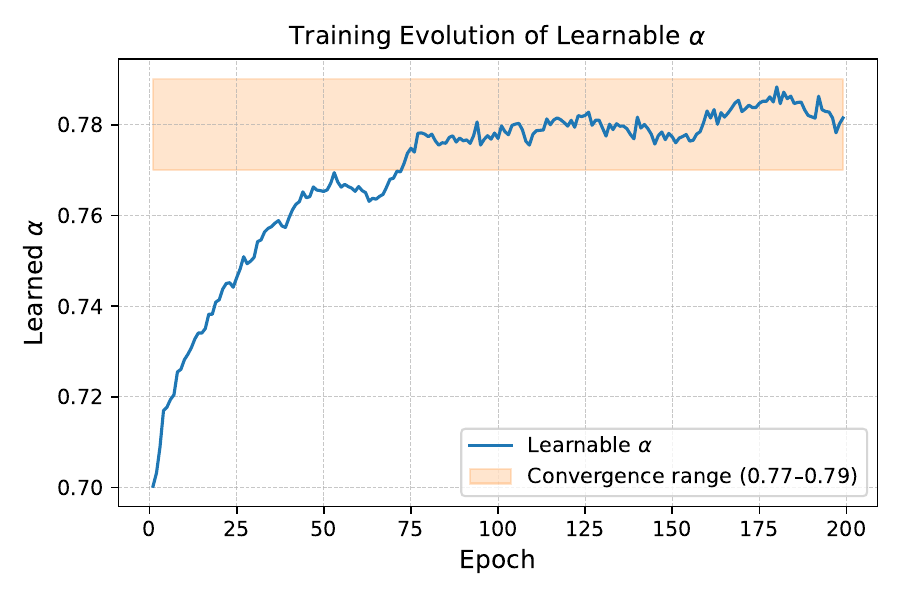}
	\caption{Training evolution of the learnable fusion coefficient $\alpha$ in GGEO over 200 epochs. 
		After an initial increasing phase, $\alpha$ fluctuates within a narrow range, indicating a stable fusion preference during training.}
	\label{fig:13}
\end{figure}
As shown in Table~\ref{tab:ablation_GGEO}, the single-direction configuration obtains the lowest AP and AP$_{75}$, indicating that one gradient orientation is insufficient to describe the weak and irregular boundaries of infrared gas plumes. 
When two orthogonal directions ($0^\circ$ and $90^\circ$) are used, the model achieves the best overall AP of 28.8\% and the highest AP$_{50}$ of 81.1\%. 
This suggests that horizontal and vertical gradient cues provide strong coarse boundary information for region-level plume detection.

When diagonal directions ($45^\circ$ and $135^\circ$) are further introduced, the model achieves the best AP$_{75}$ and AP$_S$, reaching 10.3\% and 24.0\%, respectively. 
This indicates that additional oblique-direction responses are helpful for stricter localization and small plume detection, where weak boundaries may appear with irregular or slanted shapes. 
However, the overall AP and AP$_{50}$ are slightly lower than those of the two-direction configuration. 
This result suggests that adding more directional responses can improve fine-scale boundary sensitivity, but may also introduce additional background edge responses in complex thermal scenes.

Overall, the ablation results show that gradient direction modeling influences different detection metrics in different ways. 
The two-direction setting provides the best overall AP and AP$_{50}$, while the four-direction setting is more beneficial for AP$_{75}$ and AP$_S$. 
In this study, the four-direction configuration is retained in MSEPM because
it provides stronger AP$_{75}$ and AP$_S$ performance, which is beneficial
for high-IoU localization and small-plume detection, while introducing
negligible additional computational cost.

\subsubsection{Effect of the Fusion Coefficient $\alpha$ in GGEO}

We conducted an ablation study to analyze the effect of the fusion coefficient $\alpha$ in the gradient--Gabor edge operator (GGEO). 
As shown in Figure~\ref{fig:12}, different settings of $\alpha$ lead to different detection results across AP, AP$_{50}$, and AP$_S$, indicating that the balance between gradient cues and Gabor-response
cues affects boundary-sensitive structural-prior representation. 
When $\alpha$ is fixed, intermediate values generally provide better small-object performance than extreme settings. 
The gradient-only setting ($\alpha=1.0$) achieves relatively strong overall performance, whereas the Gabor-only setting ($\alpha=0$) gives lower accuracy, especially for small plume targets. 
This suggests that gradient cues are important for local boundary sensitivity, while gabor-response cues provide complementary structural information.

Compared with manually fixed values, the learnable $\alpha$ achieves the best overall performance among the tested settings in terms of AP, AP$_{50}$, and AP$_S$. 
This result indicates that allowing the network to adjust the relative contribution of directional-gradient and Gabor-response cues during training is more effective than manually selecting a fixed fusion coefficient. 
It also reduces the need for empirical tuning of $\alpha$ across different datasets or training settings.

We further tracked the evolution of the learnable $\alpha$ during training, as shown in Figure~\ref{fig:13}. 
The coefficient is initialized at 0.7 and increases during the early training stage. 
After approximately 50 epochs, it enters a relatively stable range of about 0.77--0.79 and remains within this narrow interval for the rest of training. 
This trend suggests that the network does not collapse to either a gradient-only or Gabor-only configuration. 
Instead, it converges to a fusion preference that assigns a larger weight to gradient cues while still retaining complementary Gabor-response information. 
Therefore, the learnable fusion coefficient provides a stable and data-driven way to combine the two edge cues.

\subsubsection{Edge-Prior Replacement and Operator Comparison}

\begin{table}[t]
	\centering
	\caption{Comparison of edge operators and controlled edge-prior replacements
		within MSEPM on the IIG dataset.}
	\label{tab:operators_comparison}
	\small
	
	\textbf{(a) Classical edge-operator comparison}\\[2pt]
	
	\begin{tabular}{lcccccccc}
		\toprule
		Operator & AP & AP$_{50}$ & AP$_{75}$ & AP$_S$ & AP$_M$ & AP$_L$ & GFLOPs & Params \\
		\midrule
		Sobel~\citep{heath1998comparison}
		& \textbf{27.9} & 79.7 & 8.1 & 21.5 & \textbf{32.9} & 33.6 & \textbf{59.4} & \textbf{21.109} \\
		Canny~\citep{canny1986computational}
		& 25.8 & 80.0 & 7.2 & 22.1 & 27.7 & 31.2 & 60.1 & 21.110 \\
		Laplacian~\citep{wang2007laplacian}
		& 26.7 & 79.4 & 8.2 & 21.9 & 30.1 & 29.7 & 59.8 & 21.110 \\
		GGEO
		& \textbf{27.9} & \textbf{80.4} & \textbf{10.3} & \textbf{24.0} & 30.6 & \textbf{38.1} & 59.8 & 21.119 \\
		\bottomrule
	\end{tabular}
	
	\vspace{5pt}
	
	\textbf{(b) Edge-prior randomization and replacement}\\[2pt]
	
	\begin{tabular}{lcccccccc}
		\toprule
		Prior source & AP & AP$_{50}$ & AP$_{75}$ & AP$_S$ & AP$_M$ & AP$_L$  & GFLOPs & Params \\
		\midrule
		No edge prior (baseline)
		& 26.8 & 77.8 & 8.7 & 19.9 & 32.0 & 37.1 &\textbf{56.9}  &19.873 \\
		Statistics-matched random prior
		& 26.0 & 78.8 & 6.0 & 21.8 & 29.3 & 31.0 &60.0  &21.234 \\
		Sobel-only
		& 27.9 & 79.7 & 8.1 & 21.5 & \textbf{32.9} & 33.6 &59.4  &\textbf{21.109} \\
		Pretrained HED~\citep{xie2015holistically}
		& \textbf{28.3} & \textbf{82.3} & 8.7 & 23.1 & 31.0 & 36.2  &122.6  &35.947\\
		GGEO-based MSEPM
		& 27.9 & 80.4 & \textbf{10.3} & \textbf{24.0} & 30.6 & \textbf{38.1}  &59.8  &21.119\\
		\bottomrule
	\end{tabular}
\end{table}

Table~\ref{tab:operators_comparison} compares the proposed gradient--Gabor edge operator (GGEO) with several classical edge operators when they are embedded in the same multi-scale edge perception module (MSEPM). 
For a fair comparison, only the edge operator is replaced, while the remaining network architecture, training settings, and feature fusion strategy are kept unchanged.

Sobel achieves the same overall AP as GGEO and requires slightly lower computational cost. 
This indicates that simple gradient-based edge priors are already effective for infrared gas plume detection. 
However, GGEO obtains higher AP$_{50}$, AP$_{75}$, AP$_S$, and AP$_L$ than Sobel. 
In particular, AP$_{75}$ increases from 8.1\% to 10.3\%, AP$_S$ increases from 21.5\% to 24.0\%, and AP$_L$ increases from 33.6\% to 38.1\%. 
These results suggest that combining directional-gradient cues with
Gabor-response information is beneficial for stricter localization,
small-plume detection, and large diffuse plume representation.

Canny and Laplacian achieve lower overall AP than Sobel and GGEO under the same setting. 
This may be because their edge responses are more sensitive to thresholding, noise, or second-order intensity variations in low-contrast infrared scenes. 
Compared with these classical operators, GGEO provides a more balanced edge representation by combining local gradient sensitivity with gabor-response cues. 
Nevertheless, the improvement in overall AP over Sobel is not observed, and GGEO introduces a small amount of additional computation. 
Therefore, the advantage of GGEO should be interpreted mainly in terms of AP$_{50}$, AP$_{75}$, AP$_S$, and AP$_L$, rather than as a uniform improvement across all metrics.

{The Sobel--GGEO comparison isolates the choice of edge operator within the
	same MSEPM pathway, with all other components kept unchanged. Under this
	setting, GGEO maintains the same overall AP of 27.9\% as Sobel while
	improving AP$_{50}$ from 79.7\% to 80.4\%, AP$_{75}$ from 8.1\% to
	10.3\%, AP$_S$ from 21.5\% to 24.0\%, and AP$_L$ from 33.6\% to
	38.1\%; AP$_M$ decreases from 32.9\% to 30.6\%. The computational cost
	increases from 59.4 to 59.8 GFLOPs, corresponding to an operator-level
	overhead of 0.4 GFLOPs. Thus, Sobel provides the lower-cost configuration
	when overall AP and computational efficiency are prioritized, whereas GGEO
	favors stricter localization and small- or large-plume performance at a
	modest additional cost. The approximately 2.9-GFLOP overhead of MSEPM refers
	to introducing the complete multi-scale structural-prior pathway, whereas
	the additional cost of GGEO relative to Sobel within that pathway is only
	0.4 GFLOPs.}

{To further determine whether the benefit of MSEPM originates from meaningful
	boundary-sensitive structural priors rather than from generic feature
	perturbation, we conducted the controlled prior-randomization and replacement
	experiment summarized in Table~\ref{tab:operators_comparison}(b). In addition to the original GGEO-based MSEPM, the structural prior was
	replaced by a statistics-matched random prior, Sobel-only gradient
	responses, and a pretrained HED edge prior.
	For the random-prior control, we first computed the GGEO output
	$\mathbf{E}_0$ without gradient tracking and generated a Gaussian random
	tensor of the same shape. The random tensor was normalized and then
	rescaled channel-wise to match the spatial mean and standard deviation of
	the corresponding GGEO response. This statistics-matched construction
	reduces differences caused solely by feature scale or first- and
	second-order response statistics.
	For the HED replacement, the raw augmented input image was resized to
	$320\times320$ and processed by a pretrained frozen HED network without
	gradient propagation. The resulting single-channel edge map was resized
	to the spatial resolution of the corresponding P2 feature, replicated
	across channels, concatenated with the original P2 feature, and projected
	through a $1\times1$ convolution to obtain the initial structural prior.
	For all replacement variants, the subsequent progressive max-pooling and
	scale-specific projection pathway of MSEPM was kept unchanged.
	A no-edge-prior baseline is also included to determine whether random
	perturbation itself provides a consistent detection benefit. All variants were evaluated under the same dataset split, training schedule,
	detector input resolution, and detection settings.	
	The statistics-matched random prior achieves an AP of 26.0\% and an
	AP$_{75}$ of 6.0\%, both lower than the no-edge-prior baseline values
	of 26.8\% and 8.7\%, respectively. Although its AP$_S$ increases from 19.9\% to 21.8\%, the
	degradation in overall AP, strict-localization performance, AP$_M$, and
	AP$_L$ indicates that random feature perturbation does not provide a
	consistent detection advantage. This result suggests that the contribution
	of MSEPM cannot be explained solely by increased feature diversity.
	Meaningful structural priors provide more consistent detection gains
	across the evaluated metrics.
	The Sobel-only variant achieves an AP of 27.9\% and an AP$_S$ of 21.5\%,
	while the pretrained HED prior obtains the highest overall AP of 28.3\%
	and an AP$_S$ of 23.1\%. The GGEO-based MSEPM achieves an AP of 27.9\%,
	but obtains the best AP$_{75}$ of 10.3\% and the best AP$_S$ of 24.0\%.
	Compared with the statistics-matched random prior, GGEO improves AP by 1.9 percentage
	points, AP$_{75}$ by 4.3 percentage points, and AP$_S$ by 2.2 percentage
	points.	
	These results indicate that structurally meaningful edge priors are more
	effective than random perturbation for infrared gas plume detection. At the
	same time, the pretrained HED prior achieves slightly higher overall AP than
	GGEO, so the advantage of the proposed GGEO should not be interpreted as a
	uniform improvement across all metrics. Instead, its main benefit is observed in AP$_{75}$ and AP$_S$,
	suggesting that the combination of directional-gradient and
	Gabor-response cues is particularly useful for performance at a stricter
	IoU threshold and for small-plume representation.}

\begin{table}[htbp]
	\centering
	\caption{Ablation study of CASR-PAN components on the IIG dataset.
		The first group evaluates the adaptive routing paths, while the second
		group examines the importance-estimator branches and branch-weighting
		strategy.}
	\label{tab:ablation_routing}
	\resizebox{\textwidth}{!}{
		\begin{tabular}{lcccccccc}
			\toprule
			Model & AP & AP$_{50}$ & AP$_{75}$ & AP$_S$ & AP$_M$ & AP$_L$ &
			GFLOPs (G) & Params (M) \\
			\midrule
			
			\multicolumn{9}{c}{\textit{Routing-path ablation}} \\
			\midrule
			Naive additive fusion
			& 25.9 & 75.9 & 8.5 & 23.6 & 28.0 & 38.3 & 45.8 & 16.939 \\
			
			Without deep-to-mid fusion
			& 27.5 & 79.5 & 7.6 & 22.5 & 31.1 & 36.8 & 45.8 & 16.939 \\
			
			Without deep-to-shallow fusion
			& 25.7 & 76.2 & 6.8 & 19.2 & 30.5 & 40.6 & 45.8 & 16.939 \\
			
			Without shallow-to-mid fusion
			& 25.7 & 75.8 & 6.2 & 19.8 & 29.7 & 40.2 & 45.8 & 16.939 \\
			
			Without mid-level self-fusion
			& 28.3 & \textbf{80.3} & 8.3 & 22.3 & 31.9 & \textbf{43.5}
			& 45.8 & 16.939 \\
			
			CASR-PAN
			& \textbf{29.4} & 79.1 & \textbf{12.3} & \textbf{23.0}
			& \textbf{33.6} & 42.2 & 45.8 & 16.939 \\
			
			\midrule
			\multicolumn{9}{c}{\textit{Importance-estimator ablation}} \\
			\midrule
			
			Equal-weight IE
			& 26.5 & 77.9 & 6.5 & 20.1 & 31.7 & 31.9
			& 45.8 & 16.939 \\
			
			IE without global branch
			& 26.7 & 77.1 & 8.6 & 21.0 & 30.4 & 41.8
			& 45.8 & 16.931 \\
			
			IE without local branch
			& 27.5 & \textbf{79.7} & 7.9 & 20.6 & 31.8 & 41.7
			& \textbf{45.3} & \textbf{16.898} \\
			
			IE without diversity branch
			& 27.3 & 77.9 & 8.2 & 20.1 & 31.5 & \textbf{42.5}
			& 45.7 & 16.931 \\
			
			Full IE
			& \textbf{29.4} & 79.1 & \textbf{12.3} & \textbf{23.0}
			& \textbf{33.6} & 42.2 & 45.8 & 16.939 \\
			
			\bottomrule
		\end{tabular}
	}
\end{table}

\begin{table}[htbp]
	\centering
	\caption{Sensitivity analysis of the fixed modulation coefficients $\beta$ and $\gamma$ in CASR-PAN on the IIG dataset. One coefficient is varied while the other is fixed at its default value.}
	\label{tab:casr_sensitivity}
	{
		\resizebox{0.6\textwidth}{!}{
			\begin{tabular}{lcccccccc}
				\toprule
				Analysis & $\beta$ & $\gamma$ & AP & AP$_{50}$ & AP$_{75}$ & AP$_S$ & AP$_M$ & AP$_L$ \\
				\midrule
				\multicolumn{9}{c}{\textit{$\beta$ sensitivity ($\gamma=1.0$)}} \\
				\midrule
				$\beta$ variation & 0.00 & 1.00 & 27.5 & 78.4 & 8.6 & 21.6 & 31.1 & \textbf{42.6} \\
				$\beta$ variation & 0.25 & 1.00 & 28.2 & 77.8 & 10.3 & 21.9 & 32.2 & 35.8 \\
				Default & 0.50 & 1.00 & \textbf{29.4} & \textbf{79.1} & \textbf{12.3} & \textbf{23.0} & \textbf{33.6} & 42.2 \\
				$\beta$ variation & 0.75 & 1.00 & 25.7 & 76.5 & 6.8 & 21.3 & 28.5 & 39.0 \\
				$\beta$ variation & 1.00 & 1.00 & 25.9 & 76.9 & 7.4 & 20.7 & 29.8 & 36.3 \\
				\midrule
				\multicolumn{9}{c}{\textit{$\gamma$ sensitivity ($\beta=0.5$)}} \\
				\midrule
				$\gamma$ variation & 0.50 & 0.50 & 26.7 & 78.6 & 7.0 & 20.2 & 30.6 & 39.7 \\
				Default & 0.50 & 1.00 & \textbf{29.4} & 79.1 & \textbf{12.3} & \textbf{23.0} & \textbf{33.6} & \textbf{42.2} \\
				$\gamma$ variation & 0.50 & 1.50 & 26.9 & \textbf{79.8} & 9.2 & 22.3 & 29.7 & 39.1 \\
				\bottomrule
			\end{tabular}
	}}
\end{table}

\begin{table}[htbp]
	\centering
	\caption{Comparison of different neck structures on the IIG dataset.}
	\label{tab:neck_comparison}
	\resizebox{\textwidth}{!}{
		\begin{tabular}{lccccccccc}
			\toprule
			Model & AP & AP$_{50}$ & AP$_{75}$ & AP$_S$ & AP$_M$ & AP$_L$ & GFLOPs (G) & Params (M) \\
			\midrule
			PANet~\citep{liu2018path} & 27.3 & \textbf{80.9} & 8.0 & 22.1 & 30.0 & 39.2 & 103.8 & 21.463 \\
			BiFPN~\citep{tan2020efficientdet} & 27.0 & 76.6 & 8.3 & 20.8 & 31.3 & 38.9 & 64.3 & 20.300 \\
			NAS-FPN~\citep{ghiasi2019fpn} & 25.1 & 76.4 & 7.5 & 19.3 & 29.9 & 26.1 & 93.8 & 19.756 \\
			CASR-PAN & \textbf{29.4} & 79.1 & \textbf{12.3} & \textbf{23.0} & \textbf{33.6} & \textbf{42.2} & \textbf{45.8} & \textbf{16.939} \\
			\bottomrule
		\end{tabular}
	}
\end{table}

\subsubsection{Ablation Study of CASR-PAN Components}

To further isolate the contribution of the internal components of CASR-PAN,
we conducted three complementary analyses on the IIG dataset. The first
evaluates the contribution of individual routing paths. The second examines
the global, local, and diversity branches of the importance estimator (IE),
together with the strategy used to combine these branches. The third evaluates
the sensitivity of the fixed modulation coefficients $\beta$ and $\gamma$.
All experiments use the same RT-DETR-R18 backbone, detection head, training
strategy, input resolution, dataset split, and optimization settings. Only
the examined CASR-PAN component or coefficient is changed.

As summarized in Table~\ref{tab:ablation_routing}, the routing-path variants
retain the same model scale, whereas removing individual IE branches leads
to small changes in computational cost and parameter count because the
corresponding branch operations are removed. The coefficient sensitivity
analysis in Table~\ref{tab:casr_sensitivity} uses the same controlled
CASR-PAN-only configuration and changes only $\beta$ or $\gamma$ while keeping
the other coefficient fixed at its default value. This controlled setting
allows the effects of the routing structure, internal importance-estimation
mechanism, and fixed modulation coefficients to be evaluated independently
of the proposed plume-oriented backbone.

The naive additive fusion baseline obtains an AP of 25.9\%, indicating that simple feature summation is not sufficient for effective multi-scale gas plume representation. 
Compared with this baseline, the full CASR-PAN improves AP to 29.4\%, AP$_{75}$ to 12.3\%, AP$_S$ to 23.0\%, and AP$_M$ to 33.6\%. 
{These improvements suggest that content-dependent routing helps the neck assign larger contributions to informative plume-related features while limiting the contribution of less informative cross-scale features during aggregation.}

Removing individual routing paths leads to different performance changes across scale-specific metrics. 
When the deep-to-mid fusion path is removed, AP decreases from 29.4\% to 27.5\%, and AP$_{75}$ decreases from 12.3\% to 7.6\%, suggesting that semantic information from deeper features is important for accurate plume localization. 
Removing the deep-to-shallow fusion path causes a clear decrease in AP$_S$, from 23.0\% to 19.2\%, indicating that high-level semantic guidance is useful for small plume detection. 
Similarly, removing the shallow-to-mid fusion path reduces AP$_S$ and AP$_M$, showing that low-level spatial details contribute to mid-level feature refinement.

The variant without mid-level self-fusion achieves a slightly higher AP$_{50}$ and AP$_L$ than the full CASR-PAN, but its AP and AP$_{75}$ are lower. 
This indicates that self-fusion is more beneficial for overall detection accuracy and stricter localization, although it may not always maximize every scale-specific metric. 
Overall, the full CASR-PAN achieves the best AP, AP$_{75}$, AP$_S$, and AP$_M$ among the tested variants. 
These results indicate that combining cross-scale routing with self-enhancement provides a more effective feature aggregation strategy for weak and scale-varying infrared gas plumes.

{
	We further examined the internal importance estimator to determine whether
	its global, local, and diversity branches provide complementary information.
	Starting from the full IE configuration, each branch was removed
	individually while the remaining branches and all other network components
	were kept unchanged. Removing the global branch decreases AP from 29.4\%
	to 26.7\%, while removing the local and diversity branches reduces AP to
	27.5\% and 27.3\%, respectively. Thus, removing any one of the three
	branches leads to a reduction in overall detection accuracy.
}

{
	The effect is more pronounced for stricter localization and small-plume
	detection. Relative to the full IE, whose AP$_{75}$ and AP$_S$ are
	12.3\% and 23.0\%, respectively, removing the global branch reduces these
	metrics to 8.6\% and 21.0\%. Removing the local branch results in
	7.9\% AP$_{75}$ and 20.6\% AP$_S$, while removing the diversity branch
	results in 8.2\% AP$_{75}$ and 20.1\% AP$_S$. These consistent decreases
	indicate that global context, localized spatial responses, and
	feature-diversity cues provide complementary information for importance
	estimation, particularly for strict localization and small weak-plume
	representation.
}

{
	We also compared the proposed learnable branch-weighting mechanism with an
	equal-weight variant in which the global, local, and diversity cues are
	retained but contribute equally to the importance estimate. Equal weighting
	reduces AP from 29.4\% to 26.5\%, AP$_{75}$ from 12.3\% to 6.5\%, and
	AP$_L$ from 42.2\% to 31.9\%, despite having essentially the same
	computational cost and parameter count as the full IE. This result indicates
	that the improvement does not arise merely from combining three parallel
	importance cues; adaptively learning their relative contributions is also
	important for effective content-dependent routing.
}

{
	The full IE does not maximize every individual metric. For example, the
	variant without the local branch obtains a slightly higher AP$_{50}$ of
	79.7\%, compared with 79.1\% for the full IE, while the variant without
	the diversity branch obtains a slightly higher AP$_L$ of 42.5\%, compared
	with 42.2\%. Therefore, the three-branch estimator should not be interpreted
	as uniformly superior for every scale-specific metric. Rather, it provides
	the strongest overall balance, achieving the highest AP, AP$_{75}$,
	AP$_S$, and AP$_M$ among the evaluated IE configurations. Together, these
	results support the complementary roles of the three importance cues and
	the use of learnable weighting in the proposed content-adaptive routing
	mechanism.
}

{
	Finally, we evaluated the sensitivity of the fixed modulation coefficients
	$\beta$ and $\gamma$ in the same controlled CASR-PAN-only setting, as shown
	in Table~\ref{tab:casr_sensitivity}. For the $\beta$ analysis, $\gamma$ was
	fixed at 1.0 and $\beta$ was varied over $\{0,0.25,0.5,0.75,1.0\}$. The
	overall AP increases from 27.5\% at $\beta=0$ to 28.2\% at $\beta=0.25$
	and reaches 29.4\% at $\beta=0.5$. AP$_{75}$ follows the same trend,
	increasing from 8.6\% to 10.3\% and then to 12.3\%. Increasing $\beta$
	further to 0.75 or 1.0 reduces AP to 25.7\% and 25.9\%, respectively,
	with corresponding AP$_{75}$ values of 6.8\% and 7.4\%. These results
	indicate that a moderate positive modulation offset provides a more
	favorable balance than either removing the offset or applying a stronger
	bias. Although $\beta=0$ obtains a slightly higher AP$_L$ of 42.6\%
	than the 42.2\% obtained at $\beta=0.5$, the default $\beta=0.5$ setting
	achieves the best AP, AP$_{50}$, AP$_{75}$, AP$_S$, and AP$_M$ among the
	evaluated $\beta$ values.
}

{
	For the $\gamma$ analysis, $\beta$ was fixed at 0.5 and $\gamma$ was varied
	over $\{0.5,1.0,1.5\}$. The default $\gamma=1.0$ setting achieves 29.4\%
	AP and 12.3\% AP$_{75}$, compared with 26.7\%/7.0\% for $\gamma=0.5$
	and 26.9\%/9.2\% for $\gamma=1.5$. It also provides the highest AP$_S$,
	AP$_M$, and AP$_L$ among the evaluated $\gamma$ settings. The
	$\gamma=1.5$ variant obtains a slightly higher AP$_{50}$ of 79.8\% than
	the 79.1\% obtained at the default setting. Therefore, $\gamma=1.0$ should
	not be interpreted as uniformly maximizing every individual metric; rather,
	it provides the strongest overall balance in the tested range. Taken
	together, the sensitivity analysis supports the empirical selection of
	$\beta=0.5$ and $\gamma=1.0$ and shows that both coefficients have a
	measurable influence on the balance between content-dependent modulation
	and identity-preserved feature propagation. These conclusions are limited
	to the evaluated coefficient ranges and the controlled IIG setting.
}

\subsubsection{Comparison of CASR-PAN with Multi-Scale Fusion Strategies}

To further evaluate the effectiveness of CASR-PAN for cross-scale feature aggregation, we compared it with several representative neck architectures, including PANet~\citep{liu2018path}, BiFPN~\citep{tan2020efficientdet}, and NAS-FPN~\citep{ghiasi2019fpn}. 
All neck structures are integrated into the same backbone and detection head to ensure a fair comparison. 
The results are summarized in Table~\ref{tab:neck_comparison}.

CASR-PAN achieves the highest AP of 29.4\%, outperforming PANet, BiFPN, and NAS-FPN by 2.1, 2.4, and 4.3 percentage points, respectively. 
It also obtains the best AP$_{75}$, AP$_S$, AP$_M$, and AP$_L$, indicating that the proposed routing-based fusion is beneficial for stricter localization and multi-scale plume representation. 
In particular, the improvement in AP$_S$ suggests that CASR-PAN can better preserve weak small-plume features during feature aggregation, while the improvement in AP$_L$ indicates its effectiveness for large and diffuse plume regions.

In terms of computational cost, CASR-PAN requires 45.8 GFLOPs and 16.939 M parameters, which are lower than those of PANet, BiFPN, and NAS-FPN in this setting. 
This shows that CASR-PAN improves feature fusion effectiveness while reducing redundant computation in the neck. 
However, PANet achieves the highest AP$_{50}$ of 80.9\%, whereas CASR-PAN obtains 79.1\%. 
This suggests that PANet remains competitive for coarse region-level detection, while CASR-PAN provides better overall AP, stricter localization performance, scale-specific detection, and computational efficiency.

Overall, the comparison demonstrates that CASR-PAN provides a favorable balance between detection accuracy and computational cost for infrared gas plume detection. 
Its advantage mainly comes from content-adaptive routing, which allows the network to selectively propagate informative features across scales rather than relying on fixed fusion pathways.

\subsubsection{Comparison of Global-Context Modeling Strategies}
{
	To examine whether the effectiveness of the local--global plume block
	depends specifically on its frequency-domain global branch, we conducted
	a controlled replacement experiment using three representative
	spatial-domain global-context modeling strategies. Specifically, the
	proposed DCT-based frequency branch was replaced by a $31\times31$
	large-kernel depthwise-separable convolution~\citep{ding2022scaling}, a Non-local module~\citep{wang2018nonlocal}, and
	Swin shifted-window self-attention~\citep{liu2021swin}. For all variants, only the global-context branch was replaced, while the
	local branch, MSEPM-based structural-prior fusion, feature-fusion structure,
	backbone placement, detection head, dataset split, input resolution,
	optimizer, and training schedule were kept unchanged. The results are summarized in Table~\ref{tab:global_context_comparison}.
}

\begin{table}[htbp]
	\centering
	\caption{Comparison of alternative global-context modeling strategies
		within the local--global plume block on the IIG dataset.}
	\label{tab:global_context_comparison}
	\resizebox{\textwidth}{!}{
		\begin{tabular}{lcccccccc}
			\toprule
			Global-context branch &
			AP &
			AP$_{50}$ &
			AP$_{75}$ &
			AP$_S$ &
			AP$_M$ &
			AP$_L$ &
			GFLOPs (G) &
			Params (M) \\
			\midrule
			$31\times31$ large-kernel convolution
			& 25.3 & 74.5 & 7.3 & 17.8 & 31.2 & 31.6 & 55.1 & 18.095 \\
			
			Non-local module
			& 26.6 & 77.9 & 8.6 & 20.3 & 31.1 & 33.9 & 53.6 & 17.274 \\
			
			Swin shifted-window attention
			& 26.8 & 76.9 & \textbf{9.6} & \textbf{22.2} & 29.6 &
			\textbf{39.3} & 72.8 & 31.743 \\
			
			Proposed DCT-based frequency branch
			& \textbf{28.5} & \textbf{80.7} & 8.8 &
			\textbf{22.2} & \textbf{32.8} & 37.3 &
			\textbf{51.9} & \textbf{15.965} \\
			\bottomrule
		\end{tabular}
	}
\end{table}
{
	As shown in Table~\ref{tab:global_context_comparison}, the proposed
	DCT-based frequency branch achieves the highest overall AP of 28.5\%
	and AP$_{50}$ of 80.7\%. Compared with the $31\times31$ large-kernel
	convolution, Non-local module, and Swin attention, it improves AP by
	3.2, 1.9, and 1.7 percentage points, respectively. The DCT-based branch
	also achieves the highest AP$_M$ of 32.8\% and obtains an AP$_S$ of
	22.2\%, matching the Swin variant and exceeding the large-kernel and
	Non-local alternatives.
}
{
	The comparison also shows that the DCT-based branch is not uniformly
	superior across all metrics. Swin attention achieves a higher AP$_{75}$
	of 9.6\% compared with 8.8\% for the DCT-based branch, and a higher
	AP$_L$ of 39.3\% compared with 37.3\%. These results indicate that
	different global-context mechanisms exhibit different scale- and
	IoU-dependent behaviors, and the advantage of the proposed frequency-domain
	branch should therefore not be interpreted as universal superiority in
	representational ability.
}
{
	In terms of computational efficiency, the proposed frequency-domain branch
	requires 51.9 GFLOPs and 15.965 M parameters, which are the lowest among
	the four evaluated configurations. The Non-local and large-kernel variants
	require 53.6 and 55.1 GFLOPs, respectively, while the Swin variant increases
	the computational cost to 72.8 GFLOPs and the parameter count to
	31.743 M. Compared with Swin attention, the proposed branch reduces
	computation by 20.9 GFLOPs and the parameter count by approximately
	15.78 M while improving overall AP by 1.7 percentage points and
	AP$_{50}$ by 3.8 percentage points.
}
{
	Overall, these controlled replacements show that the improvement of the
	local--global plume block cannot be reproduced simply by introducing an
	arbitrary large-receptive-field or global-interaction operator. At the same
	time, because the DCT and Swin variants achieve the same AP$_S$ and Swin
	obtains higher AP$_{75}$ and AP$_L$, the experiment does not support a
	claim that frequency-domain modeling is uniformly superior in
	representational ability. Instead, the proposed DCT-based branch provides
	the most favorable overall accuracy--complexity trade-off among the
	evaluated alternatives for the present infrared gas plume detection task,
	combining the highest overall AP and AP$_{50}$ with the lowest GFLOPs and
	parameter count.
}

\subsection{Failure Case Analysis}

Although ECAF-Det improves the detection of weak and small gas plumes, several failure cases can still be observed from the qualitative results in Figure~\ref{fig:6}. 
First, extremely small or distant gas plumes remain difficult to detect because their spatial resolution is limited and their contrast against the thermal background is weak. 
Second, in cluttered industrial scenes, thermal structures such as pipes, valves, and high-temperature background regions may produce responses similar to plume-like patterns, leading to false positives or incomplete localization. 
Third, because gas plumes are semi-transparent and gradually fade into the surrounding background, their boundaries are often ambiguous. 
In such cases, the predicted bounding box may cover the high-contrast core of the plume but fail to include weak peripheral regions, which partly explains the relatively low AP$_{75}$ compared with AP$_{50}$. 
{Finally, high-IoU evaluation may be sensitive to small variations in the
	annotated spatial extent of diffuse plume regions. Since independent
	multi-annotator bounding-box annotations are not available, inter-annotator
	agreement is not quantified in the present study.}
{The matched-box statistics in Section~4.3 further indicate that systematic
	box enlargement is not the dominant localization error. Instead, the
	remaining high-IoU error is associated with imperfect plume-extent coverage
	and residual box alignment. Thus, plume-boundary ambiguity remains a
	contributing factor, but it should not be regarded as the sole explanation
	for the AP$_{50}$--AP$_{75}$ gap.}

These observations indicate that the remaining errors are mainly associated with small target size, low contrast, background thermal clutter, and boundary ambiguity. 
Future work will therefore focus on improving high-IoU localization, incorporating temporal plume information from video sequences, and developing uncertainty-aware annotation or evaluation strategies for diffuse gas plume boundaries.

\section{Conclusion}

This study presents ECAF-Det, an edge-aware and content-adaptive feature fusion detector for infrared gas leak detection in industrial monitoring scenarios. 
The proposed framework is designed to address the weak contrast, semi-transparency, diffuse boundaries, and scale variation of infrared gas plumes. 
A plume-oriented local--global backbone is introduced to preserve fine boundary cues while capturing long-range contextual continuity of diffuse plume regions. 
{
	The multi-scale edge perception module (MSEPM) provides hierarchical
	boundary-sensitive structural priors based on gradient and Gabor-response cues,
	whose task relevance is optimized through the end-to-end detection objective rather
	than through explicit plume-edge supervision.
} 
{In addition, the content-adaptive sparse routing path aggregation network (CASR-PAN) adaptively regulates cross-scale feature propagation and limits the contribution of less informative responses during multi-scale fusion.}

Experiments on the IIG and LangGas datasets demonstrate that ECAF-Det achieves favorable detection performance among the evaluated detectors. 
On the IIG dataset, ECAF-Det obtains an AP of 29.8\%, an AP$_{50}$ of 84.3\%, and an AP$_S$ of 25.3\%, improving the RT-DETR-R18 baseline by 3.0, 6.5, and 5.4 percentage points, respectively. 
On the LangGas dataset, ECAF-Det achieves an AP of 36.3\% and an AP$_{50}$ of 68.5\%. 
{
	Ablation studies further show that the plume-oriented backbone,
	the task-adaptive integration of boundary-sensitive structural priors,
	and content-adaptive routing contribute to weak-plume representation and
	multi-scale feature aggregation.
} 
The model requires 43.7 GFLOPs and 14.28 M parameters, indicating a favorable balance between detection accuracy and computational cost under the current experimental setting.

{Despite these improvements, several limitations remain. First, the current
	study focuses on bounding-box-based gas plume detection and does not estimate
	gas concentration, leakage rate, or source emission strength. Controlled per-image metadata for physical imaging distance and leakage rate
	are also unavailable in the IIG dataset, preventing stratified evaluation
	with respect to these variables. Second,
	extremely small, distant, or visually indistinct gas plumes remain difficult
	to localize accurately, especially under cluttered thermal backgrounds.
	Third, the current evaluation is limited to the adopted within-dataset
	training/test splits of IIG and LangGas. Although the results demonstrate
	consistent performance improvements on both datasets, broader
	generalization across independent industrial sites, infrared cameras,
	gas types, and environmental conditions has not yet been established.
	The separate evaluations on IIG and LangGas should therefore not be
	interpreted as evidence of real-to-real cross-site or synthetic-to-real
	generalization. Further validation using data from different industrial
	sites, infrared cameras, gas types, and environmental conditions is needed
	to assess broader applicability.}
{Future work will therefore focus on cross-site infrared gas leak detection, video-based temporal plume modeling, multi-gas detection, and systematic deployment evaluation and optimization on low-power industrial edge devices.}
In addition, adapting or distilling vision foundation models for infrared gas monitoring may help reduce data dependence and improve transferability under limited annotation conditions.

\section*{Acknowledgements}
Not applicable.

\section*{Funding}
This work was supported in part by the National Natural Science Foundation of China (No. 62303368), by the National Foreign Experts Project (No. H20251091), and by the Qin Chuang Yuan ``Scientist + Engineer'' Team Construction Project (No. 2024QCY-KXJ-172).

\section*{CRediT Authorship Contribution Statement}
Dongsheng Li: Conceptualization, Data curation, Formal analysis, Investigation, Methodology, Software, Validation, Visualization, Writing -- original draft, Writing -- review \& editing. Tianli Ma: Funding acquisition, Supervision. Siling Wang: Software, Validation, Visualization, Writing -- original draft. Beibei Duan: Visualization. Song Gao: Funding acquisition, Supervision, Writing -- review \& editing.

\section*{Declaration of Competing Interest}
The authors declare that they have no known competing financial interests or personal relationships that could have appeared to influence the work reported in this paper.

\section*{Data Availability}

Data will be made available on request.

\section{List of Abbreviations}

\begingroup
\scriptsize
\setlength{\tabcolsep}{3pt}
\renewcommand{\arraystretch}{1.10}

\begin{longtable}{p{0.16\textwidth}p{0.31\textwidth}p{0.16\textwidth}p{0.31\textwidth}}
	\caption{List of abbreviations and symbols used in this paper.}
	\label{tab:abbreviations}\\
	
	\toprule
	\textbf{Abbr./Symbol} & \textbf{Meaning} 
	& \textbf{Abbr./Symbol} & \textbf{Meaning} \\
	\midrule
	\endfirsthead
	
	\toprule
	\textbf{Abbr./Symbol} & \textbf{Meaning} 
	& \textbf{Abbr./Symbol} & \textbf{Meaning} \\
	\midrule
	\endhead
	
	RT-DETR & Real-Time Detection Transformer
	& ECAF-Det & Edge-Aware and Content-Adaptive Feature Fusion Detector \\
	
	GGEO & Gradient--Gabor Edge Operator
	& MSEPM & Multi-Scale Edge Perception Module \\
	
	CASR-PAN & Content-Adaptive Sparse Routing Path Aggregation Network
	& IE & Importance Estimator \\
	
	AIMM-F & Adaptive Information Modulation Module for Fusion
	& AIMM-S & Adaptive Information Modulation Module for Self-Enhancement \\
	
	FPN & Feature Pyramid Network
	& PANet & Path Aggregation Network \\
	
	BiFPN & Bi-directional Feature Pyramid Network
	& NAS-FPN & Neural Architecture Search Feature Pyramid Network \\
	
	DCT & Discrete Cosine Transform
	& IDCT & Inverse Discrete Cosine Transform \\
	
	AIFI & Attention-based Intra-scale Feature Interaction
	& ERF & Effective Receptive Field \\
	
	GFLOPs & Giga Floating-Point Operations
	& AP & Average Precision \\
	
	AP$_S$ & Average Precision for Small Objects
	& AP$_M$ & Average Precision for Medium Objects \\
	
	AP$_L$ & Average Precision for Large Objects
	& $\alpha$ & Learnable fusion coefficient in GGEO \\
	
	$\beta$ & Positive modulation offset
	& $\gamma$ & Identity-preservation coefficient \\
	
	\bottomrule
\end{longtable}

\endgroup

\section*{Declaration of Generative AI and AI-assisted Technologies in the Manuscript Preparation Process}
During the preparation of this work, the authors used ChatGPT to assist with language polishing and manuscript organization. After using this tool, the authors reviewed and edited the content as needed and take full responsibility for the content of the submitted manuscript.

\end{document}